\documentclass[sigconf]{acmart}
\usepackage[capitalize]{cleveref}
\crefname{section}{Sec.}{Secs.}
\Crefname{section}{Section}{Sections}
\Crefname{table}{Table}{Tables}
\crefname{table}{Tab.}{Tabs.}

\definecolor{lightgray}{rgb}{0.83, 0.83, 0.83}

\AtBeginDocument{%
  \providecommand\BibTeX{{%
    \normalfont B\kern-0.5em{\scshape i\kern-0.25em b}\kern-0.8em\TeX}}}

\setcopyright{acmcopyright}
\copyrightyear{2022}
\acmYear{2022}
\acmDOI{XXXXXXX.XXXXXXX}

\acmConference[MM '22]{ACM Multimedia '22}{October 10--17,  2022}{Lisbon, Portugal}
\acmPrice{15.00}
\acmISBN{978-1-4503-XXXX-X/18/06}
\usepackage{multirow}
\usepackage{amsmath}
\usepackage{graphicx}
\usepackage{amsmath}
\usepackage{booktabs}
\usepackage{multirow}
\usepackage{soul}
\usepackage{bbm}
\usepackage{expl3}
\ExplSyntaxOn
\newcommand\latinabbrev[1]{
	\peek_meaning:NTF . {
		#1\@}%
	{ \peek_catcode:NTF a {
			#1.\@ }%
		{#1.\@}}} 
\ExplSyntaxOff
\def\eg{\latinabbrev{e.g}}

\def\ie{\latinabbrev{i.e}}
 
\acmSubmissionID{345}

\begin{document}
\newcommand{\todo}[1]{\textcolor{orange}{(@TODO: #1)}}
\newcommand{\del}[1]{\sout{\textcolor{purple}{#1}}}
\newcommand{\note}[1]{\textcolor{orange}{#1}}
\newcommand{\rev}[1]{\textcolor{purple}{#1}}
\newcommand{\highlight}[1]{\textit{\textbf{#1}}}
\newcommand{\fig}[1]{Fig.~#1}

\title[Towards Robust Video Object Segmentation with Adaptive Object Calibration]{Towards Robust Video Object Segmentation \\with Adaptive Object Calibration}






\author{Xiaohao Xu}\authornote{The work was done when Xiaohao Xu was an intern at MSRA.}
\affiliation{\institution{Huazhong University of Science \& Technology}\city{}\country{}}
\email{xxh11102019@outlook.com}
\author{Jinglu Wang}
\affiliation{\institution{Microsoft Research Asia}\city{}\country{}}
\email{jinglwa@microsoft.com}
\author{Xiang Ming}
\affiliation{\institution{Microsoft Research Asia}\city{}\country{}}
\email{xiangming@microsoft.com}
\author{Yan Lu}
\affiliation{\institution{Microsoft Research Asia}\city{}\country{}}
\email{yanlu@microsoft.com}






\renewcommand{\shortauthors}{Xu et al.}


\begin{abstract}
In the booming video era, video segmentation attracts increasing research attention in the multimedia community.
Semi-supervised video object segmentation (VOS) aims at segmenting objects in all \textit{target} frames of a video, given annotated object masks of \textit{reference} frames. Most existing methods build pixel-wise reference-target correlations and then perform pixel-wise tracking to obtain target masks. Due to neglecting object-level cues, pixel-level approaches make the tracking vulnerable to perturbations, and even indiscriminate among similar objects.
Towards robust VOS, the key insight is to calibrate the representation and mask of each specific object to be expressive and discriminative. Accordingly, we propose a new deep network, which can adaptively construct object representations and calibrate object masks to achieve stronger robustness.
First, we construct the object representations by applying an adaptive object proxy (AOP) aggregation method, where the proxies represent arbitrary-shaped segments at multi-levels for reference. 
Then, prototype masks are initially generated from the reference-target correlations based on AOP.
Afterwards, such proto-masks are further calibrated through network modulation, conditioning on the object proxy representations.
We consolidate this conditional mask calibration process in a progressive manner, where the object representations and proto-masks evolve to be discriminative iteratively.
Extensive experiments are conducted on the standard VOS benchmarks, YouTube-VOS-18/19 and DAVIS-17. 
Our model achieves the state-of-the-art performance among existing published works, and also exhibits superior robustness against perturbations.
\end{abstract}





\maketitle
\begin{figure}[ht]
	\centering
	\includegraphics[width=0.48\textwidth]{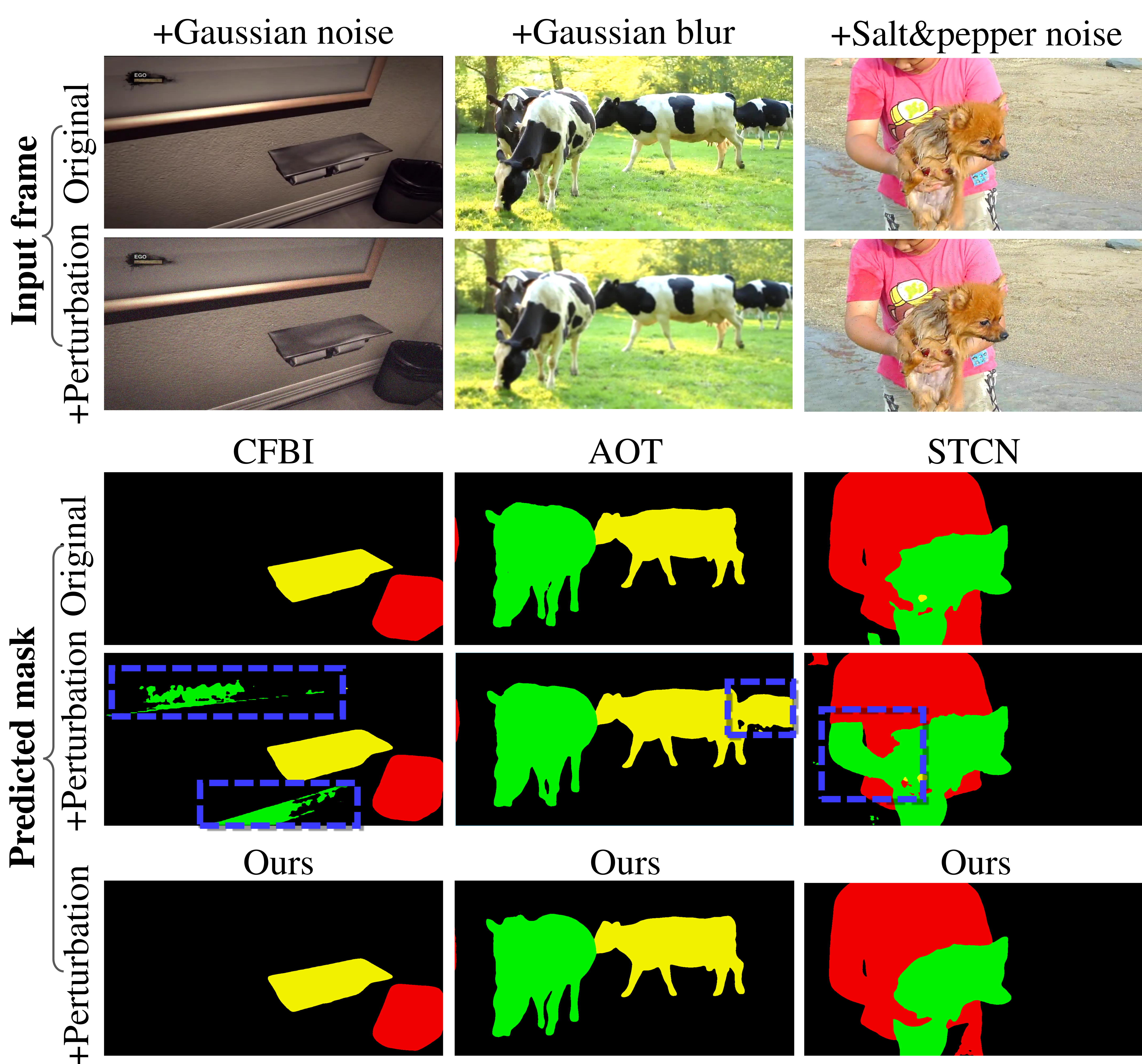}
\caption{Existing advanced VOS models, including CFBI \cite{yang2020collaborative}, AOT \cite{yang2021associating}, and STCN \cite{cheng2021rethinking}, are fragile to natural perturbations, especially for the scene containing multiple objects. Our model with adaptive object calibration shows superior robustness against perturbations.}
\label{fig:teaser}   
\end{figure}




\section{Introduction}
\label{sec:introduction}

Video Object Segmentation (VOS) is one of the most attractive video-related research problems in the multimedia area. This work focuses on the semi-supervised setting, which aims to segment one or multiple objects in a video sequence given annotated object masks of reference frames. 
In real-world videos, perturbations commonly exist due to signal noise, camera defocus, and fast motion. The model robustness becomes critical for real applications, especially for some safety-aware applications, such as autonomous driving.

However, existing VOS methods have seldom discussed the robustness against perturbations.
According to our pilot study, the performance of current advanced models \cite{cheng2021rethinking,yang2021associating,yang2020collaborative} degrade largely under simple image perturbations, as is shown in \fig{\ref{fig:teaser}}, especially for the scenes where multiple objects exist. 
Towards robust VOS, we consider two key factors that matter: 1) a robust object representation extracted from references for building correlations to the target frames; 2) a robust mask calibration process to produce pixel-wise classification conditioning on the referenced object representation.

For the representation of objects in VOS, most previous works  ~\cite{oh2019video,seong2020kernelized,lu2020video,cheng2021modular,wang2021swiftnet,xie2021efficient,hu2021learning,liang2020video,yang2020collaborative,hu2018videomatch,10.1145/3394171.3414035,10.1145/3394171.3413942} directly employ pixel-level features. Concretely, pixel-level matching \cite{hu2018videomatch} is utilized to track pixels across frames. Object-related information is only implicitly encoded in the pixel-level feature in terms of limited receptive fields. Such pixel-based representation is often error-prone, leading to noisy results, especially in the cases that similar objects co-exist \cite{liang2020video}. Meanwhile, following the fashion of object-tracking \cite{rt-vos}, some VOS works turn to object-level representations. For instance, objects are represented as box-bounded proposals \cite{voigtlaender2019boltvos,dmn-aoa} generated with an off-the-shelf object detector and predictions are made by correlating object proposals of the current frame with historical templates. Despite these methods can eliminate noisy predictions to some extent, fine-grained correspondences are lost, leading to inaccurate results in details.

For the mask calibration process, some recent works \cite{oh2019video,xie2021efficient} introduce network modulation \cite{huang2017arbitrary,park2019semantic,dumoulin2016learned,wang2018recovering} to conditionally decode the target object masks, which have already achieved successful results. Most methods \cite{oh2019video,xie2021efficient} consider individual objects and neglect interactions between different objects, which often fail in scenes with multiple objects. 
We only find a recent method, AOT \cite{yang2021associating}, encoding object-aware interactions using transformer-based association, which demonstrates that such kind of  interactions can contribute to a large performance gain. However, AOT could not discriminate similar objects well against perturbations (\fig{\ref{fig:teaser}}).

To achieve the robustness of VOS, we propose a deep network that adaptively calibrates the object representation and masks.
First, we introduce an adaptive object proxy representation to extract object-specific features from the reference. The new representation is constructed from multiple granularities for building robust correlations between the reference and target frames afterward. Then, prototype masks are initially generated from the calculated correlations. After that, we progressively perform the object mask calibration to refine the masks, conditioning on the learned object proxies. The mask calibration is implemented with network modulation, which performs channel-wise reweighting according to the learned conditioning weights. Different from previous methods, we also update the learned conditioning weights to discriminate each object from other co-existing ones during the mask evolving process. Thus, the object representation and mask are calibrated in an interleaving manner.

Our contributions are summarized as follows. 
\begin{itemize}
    \item We are the first to conduct a comprehensive study of the robustness of VOS models against perturbations. Towards robust VOS, we rethink the problem from the perspective of object representation and mask calibration, and propose a framework with stronger perturbation robustness. 
    \item We introduce an adaptive object proxy representation for referenced objects robustly, which reduces errors incurred by unstable pixel-level matching. 
    \item We calibrate the object masks by updating object representation and masks in an interleaving manner progressively, achieving discrimination among co-existing objects. 
\end{itemize}

Extensive experiments are conducted on the standard VOS benchmarks and our constructed pilot robustness benchmark.
Our model not only achieves the state-of-the-art results on the standard YouTube-VOS-18/19 and DAVIS benchmarks among existing published methods, but also exhibits superior robustness under perturbations.

\section{Related Work}
\noindent\textbf{Video Object Segmentation.}
Early video object segmentation methods \cite{perazzi2017learning,caelles2017one,khoreva2019lucid,cheng2017segflow,xu2018dynamic,zhang2012video} use online fine-tuning, calculate optical flow at high computational cost, or perform segmentation in a sequence-to-sequence manner.
Recent online VOS methods \cite{10.1145/3394171.3414035} aim at achieving good performance while maintaining a real-time speed, which can be divided into propagation-based and matching-based models. For propagation-based models \cite{perazzi2017learning,caelles2017one,khoreva2019lucid}, the guidance of segmentation masks from past frames are introduced during the process of mask decoding. For matching-based models \cite{chen2018blazingly,hu2018videomatch,zeng2019dmm,yang2020collaborative,lin2019agss,voigtlaender2019feelvos,wang2019ranet,liang2020waternet,hu2018videomatch,duke2021sstvos,mao2021joint,yang2021collaborative,9710441}, an embedding space is learnt for target objects. Recently, STM-based networks \cite{oh2019video,seong2020kernelized,lu2020video,cheng2021modular,wang2021swiftnet,xie2021efficient,hu2021learning,liang2020video,seong2021hierarchical} achieve impressive results with memory networks that memorize and read information from past frames. With adaptive object representation in multiple levels and object mask calibration for multi-object discrimination during mask decoding, our model can enhance model robustness besides better performance.

\begin{figure*}[t]
	\centering
	\includegraphics[width=\textwidth]{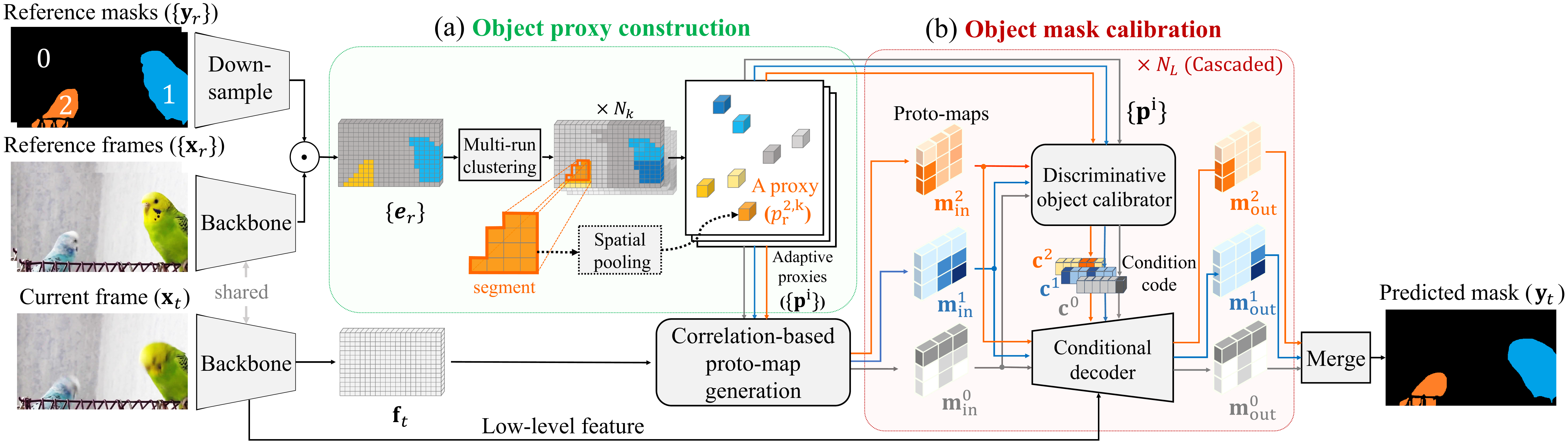} 
	\caption{{Overview of the proposed model.} Given reference frames $\{\mathbf{x}_r \}$ with annotated object masks $\{\mathbf{y}_r \}$ and the target frame $\mathbf{x}_t$, our goal is to predict the target mask $\mathbf{y}_t$ for all annotated objects ($\{0,1,2\}$ for this case). The object calibration network contains two stages, \textit{i.e.}, object proxy construction and object mask calibration.
	(a) We first construct the adaptive object-specific proxies $\mathbf{p}^i$ for each object $i$ from the reference features and masks. 
	The prototype maps $\{\mathbf{m}^i\}$ are generated with the correlation between current frame feature and adaptive proxy set $\{\mathbf{p}^i \}$. 
    (b) The object masks are progressively calibrated from $\{\mathbf{m}_{in}^i\}$ to $\{\mathbf{m}_{out}^i\}$ with condition codes $\{\mathbf{c}^i\}$. Meanwhile, the condition code for a specific object evolves to be discriminative among co-existing objects. The calibrated outputs of the last iteration are merged as the final mask of all objects $\mathbf{y}_t$. 
	}
	\label{fig:PIPELINE}
\end{figure*}

\noindent\textbf{Model Robustness.} As robustness is crucial for real-world applications with safety concerns, there is a growing trend \cite{classification-robust,detection-robust, Segmentation-robust,10.1145/3474085.3475324,hendrycks2019benchmarking,kamann2020benchmarking} to evaluate and enhance the model robustness against corruptions and perturbations. Previous study \cite{xie2019feature} suggests that adversarial perturbations on images may lead to noise in the features constructed by these networks, thus making the final prediction unstable. To reduce model fragility, many attempts have been proposed to boost the robustness of image-related tasks \cite{robust_classification, robust_HOI}.  However, there is no work to study the vulnerability of VOS models against perturbations. This work aims to fill this gap and proposes two components for the robustness enhancement of VOS models.

\section{Adaptive Object Calibration Network for Robust VOS}
We propose the adaptive object calibration network, which improves the robustness of VOS from the two key factors, \textit{i.e.}, the object representation and mask calibration.
      
\subsection{Overview.}
\label{sec:method/overview}
\fig{\ref{fig:PIPELINE}} illustrates the overview of the proposed network.
Given the target frame $\mathbf{x}_t$ and reference frames $\{\mathbf{x}_{r}\}, r \in \mathcal{S}_{ref}$ with annotated object masks $\{\mathbf{y}_{r}\}$, the goal is to predict the segmentation mask for each object $i\in\{0,1,...,N\}$ ($i$=0 indicates the background). Our basic setting uses the first and previous frame as references, namely, $\mathcal{S}_{ref}=\{1, t-1\}$.
After extracting image features with the backbone, reference features are combined with downsampled reference masks to form basic object-specific embeddings $\{ \mathbf{e}_r^i \}_{i=1}^N$. To robustly build correlations from the reference to the target frames, we introduce an adaptive proxy representation for reference object context. The proxies convey object-specific information at multiple levels, thus reducing feature matching noise. Then, initial proto-maps $\{\mathbf{m}^i\}_{i=1}^N$ of objects are generated from the calculated correlations. Afterwards, the proto-maps are progressively calibrated with condition codes $\{ \mathbf{c}^i \}$ with $N_L$ iterations. In the calibration process, each condition code $\mathbf{c}^i$ of object $i$ evolves to be discriminative from other objects and background, and then serves as channel-wise weights to modulate proto-maps from $\mathbf{m}_{in}^i$ to $\mathbf{m}_{out}^i$ with the conditional decoder. 
The final mask $\mathbf{y}_t$ is obtained by merging the output of the last conditional decoder with an {$argmax$} operation.

\subsection{Object Proxy Construction}
\label{sec:method/metric_learning}
The VOS problem is also known as mask tracking, and one of the principles is to build robust correlations from reference to target frames. The first essential step is to construct object-aware representation from the reference frames.

\noindent
\textbf{Object Proxy.}
We denote a representative embedding of each object as object proxy, \eg, $p^i \in \mathbb{R}^{1\times 1 \times C_p}$ is a proxy of object $i$. Each pixel conveys an object proxy for later pixel-wise matching between target and reference frames.

\begin{figure}[t]
	\centering
	\includegraphics[width=0.48\textwidth]{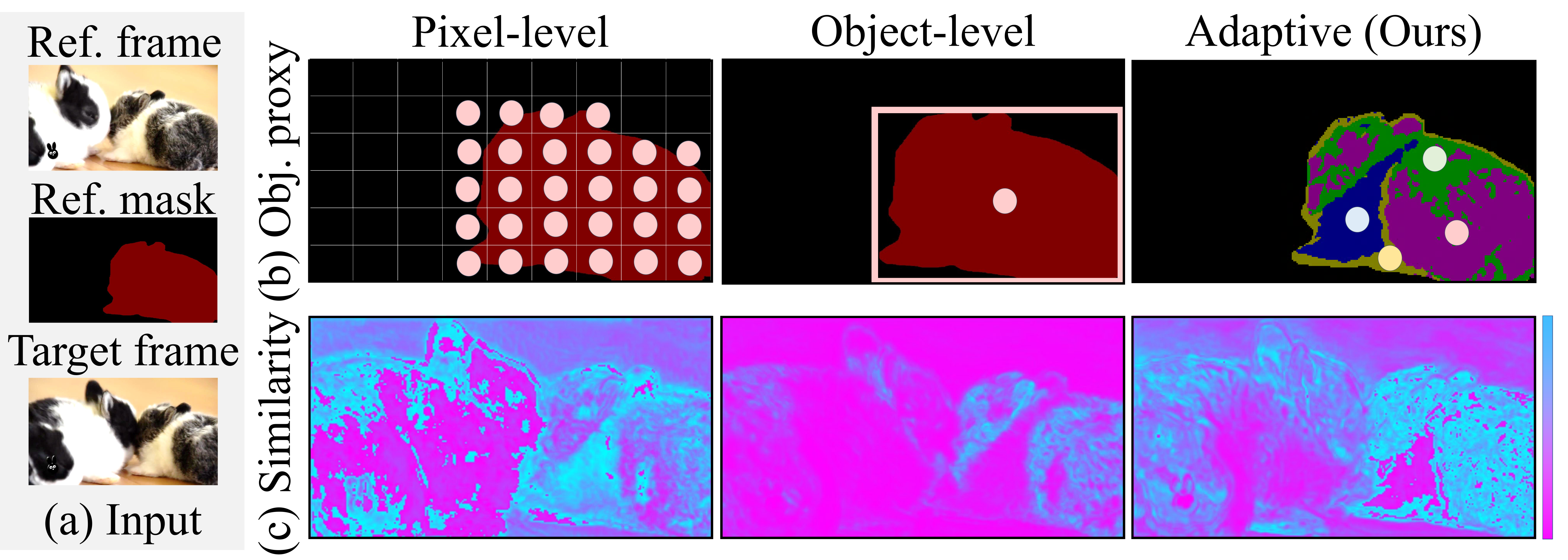}
\caption{Object proxy representation. 
(a) Given the image and object mask of the reference frame, we aim to obtain robust object representation for correlation calculation between the target and reference frames. 
(b) Object proxies are denoted as dots. For our adaptive proxy representation, we observe pixel-level embeddings can be inherently categorized into semantic clusters. Note that we only show one run of clustering here for clear visualization, while multiple runs are performed in the implementation. (c) Correlation maps in which bluer means higher similarity. Pixel-level representation is vulnerable to noise, while object-level one loses details. The proposed adaptive object representation can solve such dilemma by representing an object as a set of semantically-similar proxies via clustering-based aggregation. 
}
	\label{fig:object_representation}
\end{figure}

All previous matching-based VOS methods \cite{voigtlaender2019feelvos,yang2020collaborative,oh2019video,seong2020kernelized,lu2020video,cheng2021modular,liang2020video,hu2018videomatch,duke2021sstvos,mao2021joint} employ \textbf{\textit{pixel-level}} proxies, that is, each pixel conveys the feature of itself subject to the specific object mask (column 1 in \fig{\ref{fig:object_representation}} (b)). Let us consider the proxy map $\mathbf{p}^i \in \mathbb{R}^{H\times W \times C_p}$ for an object $i$ from reference frames:
\begin{eqnarray}
    \mathbf{p}^i = [...; \mathbf{e}_{r}^i; ...], \quad r \in \mathcal{S}_{ref}, \text{  } \mathbf{e}_{r}^{i}=\mathbf{f}_{r} \odot  \mathbbm{1}^i_{r},
\end{eqnarray}
where $[\cdot;\cdot]$ denotes channel-wise concatenation, $\mathbf{e}_r^i$ denotes the basic object-specific embedding, $\mathbf{f}_{r}$ is the extracted feature map from the backbone, $\mathbbm{1}_{r}^i$ is the downsampled object-specific binary mask from $\mathbf{y}_r$, $\odot$ denotes element-wise multiplication. 
Obviously, this representation is not robust against cases where similar objects co-exist or object appearances change drastically across frames.
Besides, the effective receptive field~\cite{luo16nipus_erf} of a single pixel-level embedding could not cover large objects or backgrounds, regardless of the global context. These issues make the embedding error-prone for object-aware correlation calculation.

To represent the proxies with object-specific cues, a straightforward practice is to use the global average pooled feature of the object. However, the \textbf{\textit{object-level}} representation could lose details.

\noindent\textbf{Adaptive Object Proxy (AOP).}
\label{sec:adaptove_proxy_representation}
To address the problems with previous object representations, we propose an adaptive proxy representation, which is a learned combination of embeddings from multiple granularities. 
We embrace the observation that pixel-level embedding corresponding to semantics can be inherently grouped into meaningful clusters, thus we construct such proxies with clustering algorithms (we use $K$-Means ~\cite{likas2003global} in implementation).
We first cluster the pixel-level feature embeddings $\mathbf{e}_r^i$ and the centroid $p_r^{i,k}$ of each cluster $c_k, k=1,...,K$ serves as a part of the proxy at different level.
Clustering is performed in multiple runs with different cluster numbers $K \in \mathcal{L}_{clu}=[K_1,..., K_{N_k}]$. Thus, the proxy can represent object information at different granularities. The proxy map $\mathbf{p}_r^{i,K}$ is produced by propagating each cluster centroid over the pixels belonging to its cluster.

We construct the final adaptive proxy representation $\mathbf{p}^{i}$ of each object $i$ as the concatenated embeddings of all the object-specific adaptive proxies, namely,
\begin{equation}
    \mathbf{p}^i = [...; \mathbf{p}_{r}^{i,K}; ...], \quad r \in \mathcal{S}_{ref}, K \in \mathcal{L}_{clu}.
\end{equation}

Accordingly, representing the objects with proxies constructed with clusters from different granularities can improve the robustness against noises compared to pixel-level ones, and also preserve more semantic details compared to object-level ones.
\fig{\ref{fig:object_representation}} illustrates proxy representation with different levels. The pink dots in \fig{\ref{fig:object_representation}} (b) represents proxies. With different proxies, the calculated correlations between reference and target frames are different. While correlations from pixel-level and object-level are either noisy or over-smoothing, adaptive proxies can generate more robust correlation maps.

\begin{figure}[t]
	\centering
	\includegraphics[width=0.48\textwidth]{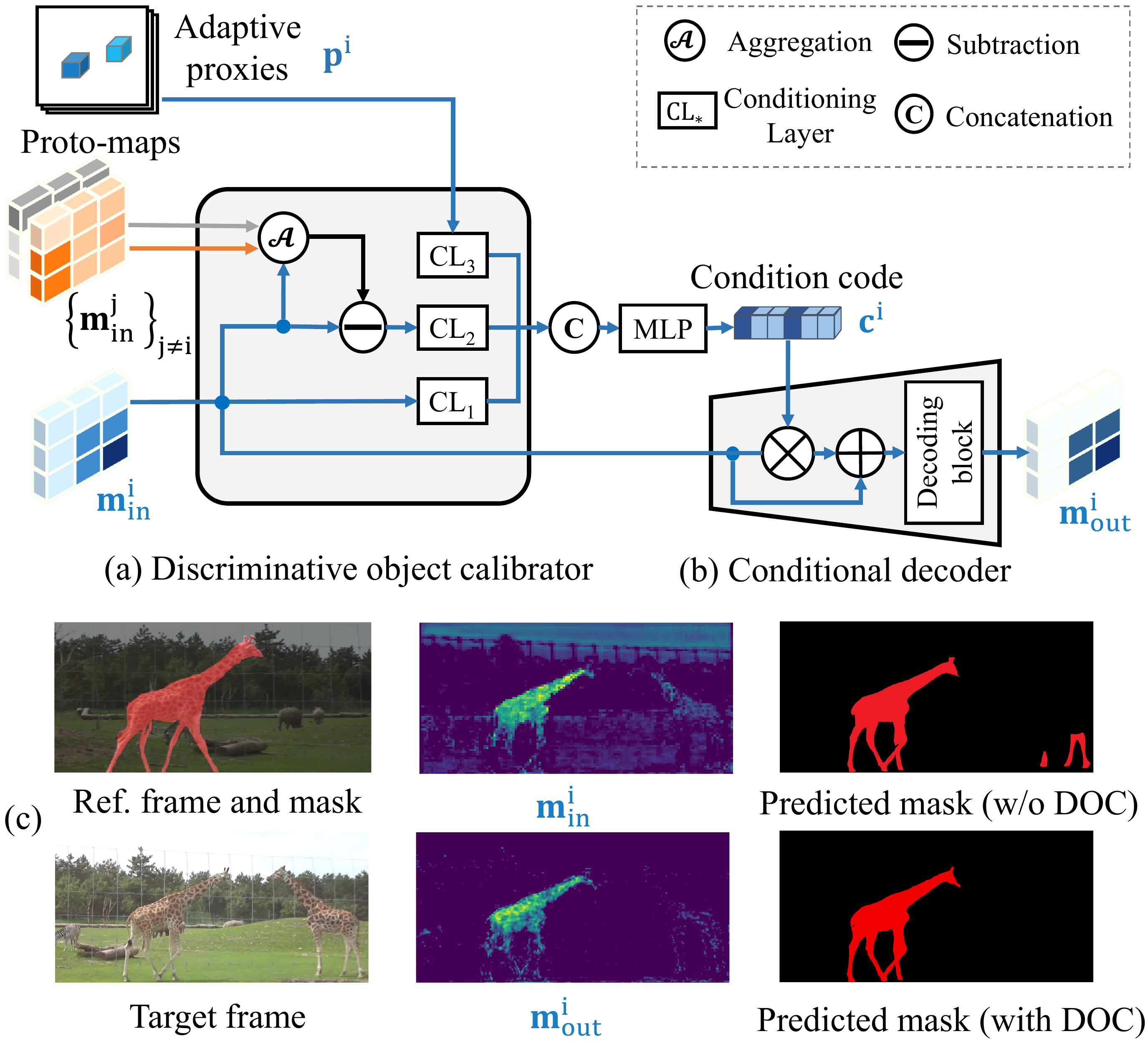}
\caption{Object mask calibration. (a) Detailed architecture of the discriminative object calibrator (DOC). (b) The condition code $\mathbf{c}^i$ with multi-object discrepancy integration is used in the conditional decoder for proto-map calibration. (c) Our mask calibration with the proposed DOC can discriminate the target giraffe from other similar ones better. }
\label{fig:multi-object-discrimination}
\end{figure}

\subsection{Object Mask Calibration}  
\label{sec:method/relational_modulation_network}

After constructing object representation from reference frames, we initialize prototype object masks (proto-map) with the correlation-based proto-map generation block. The object mask calibration process is performed progressively with $N_L$ iterations. In each iteration, the input proto-map $\mathbf{m}_{in}^i$ of object $i$ with the other $\{\mathbf{m}_{in}^{j}\}_{j\neq i}$ is first used to generate the condition code $\mathbf{c}_i$ for object $i$. $\mathbf{c}_i$ is calibrated to be discriminative from the other objects. Then, $\mathbf{m}_{in}^i$ is further calibrated with network modulation conditioning on $\mathbf{c}_i$ via the conditional decoder. The overall decoded outputs $ \{\mathbf{m}_{out}^i \}_{i=1}^N$ are merged into the final predicted mask $\mathbf{y}_t$.

\noindent\textbf{Preliminaries of Network Modulation.}
The mask calibration is implemented by network modulation, which has already been demonstrated to be effective in existing works~\cite{Yang2018osmn,yang2020collaborative,yang2021collaborative}. Network modulation is an operation to re-weight responses in different channels of a feature map, following the mechanism used in SE-Net~\cite{hu_senet_cvpr2018}, taking the form: 

\begin{eqnarray}
\label{eq:modulation}
    {z}^m_{out} = {w}^m {z}^m_{in} ,
\end{eqnarray}
where ${z}^m_{in} \in \mathbb{R}^{H \times W \times 1} $ and ${w}^m \in \mathbb{R}$ are the $m$-th channel of feature map $\mathbf{z}_{in} \in \mathbb{R}^{H \times W \times C} $ and object-specific weights $\mathbf{w}  \in \mathbb{R}^{1 \times 1 \times C}$ respectively. The modulation operation is defined as:
\begin{eqnarray}
    \mathbf{z}_{out} = \mathbf{w} \otimes \mathbf{z}_{in},
\end{eqnarray}
where $\otimes$ denotes channel-wise multiplication.

The modulation weight $\mathbf{w}$, which we call condition code, is usually aggregated as a contextual representation. The condition code is used to calculate channel-wise modulating weight for each object in the mask calibration process. 
Previous methods modulate the features $\mathbf{f}_t$ of the current frame with
features $\mathbf{f}_r$ in reference frames according to the cues (\textit{e.g.}, appearance, location) of the specific object, but object-wise interactions have hardly been considered, resulting in failure cases where visually similar objects co-exist.
We propose a discriminative object mask calibration procedure for mask modulation where information of various objects in intermediate decoding layers is exchanged with each other, thus enhancing multi-object discrimination. 

\noindent\textbf{Discriminative Object Calibration (DOC).}
\label{sec:method/relational_modulation_network/inter_object_interaction}
To calibrate object mask $\mathbf{m}^i$ to be discriminative from the other $\{ \mathbf{m}^j\}_{j \neq i}$, a prerequisite is to make the condition code $\mathbf{c}^i$ to be discriminative.
Intuitively, we enhance the discrimination of $\mathbf{c}^i$ by suppressing cues existing in the other objects, namely, by first aggregating cues from all $\{ \mathbf{m}^j\}_{j =1}^N$ and then screening out similar cues in $\mathbf{c}^i$.
The detailed operations are illustrated in \fig{\ref{fig:multi-object-discrimination}} (a). The condition code $\mathbf{c}^i$ is calibrated by taking the form:
\begin{equation}
    \mathbf{c}^{i}=MLP([CL_1(\mathbf{m}^{i}_{in}); CL_2(\mathcal{A}(\{\mathbf{m}^{j}_{in}\}_{j=1}^N)-\mathbf{m}^{i}_{in});CL_3(\mathbf{p}^{i})]),
\end{equation}
where $\mathcal{A}(\cdot)$ is an order-invariant aggregation layer for multiple inputs with a channel-wise pooling operator to aggregate object cues. Specifically, we use a channel-wise max pooling with a $1\times1$ conv in the implementation. $CL_{*}$ denotes a Conditioning Layer block, which encodes a feature map into a vector ($H\times W$ to $1 \times 1$ in the spatial dimension). Detailed implementation is introduced in Section~\ref{sec:method/relational_modulation_network/conditioning_layer}. In the discriminative object calibrator, we utilize three conditioning layer blocks $CL_1$, $CL_2$ and $CL_3$ to aggregate object information from the target object $\mathbf{m}^i$, the other objects in the current frame $\mathcal{A}(\{ \mathbf{m}^j\}_{j=1}^N)-\mathbf{m}^i$ and target object in reference frames $\mathbf{p}^i$ respectively.

Our motivation for this procedure is to incorporate discrepancy between different objects into object representation, thus reducing ambiguities between objects. 

\noindent\textbf{Conditional Decoder.}
\label{sec:method/relational_modulation_network/intra_object_interaction}
We then utilize the more discriminative condition code $\mathbf{c}^i$ to calibrate the object mask $\mathbf{m}^i$ through the conditional decoder $\theta_{dec}$.
We build $\theta_{dec}$ based on a modulation block adopted in \cite{yang2020collaborative,yang2020gated} with an additional residual block. 
As is shown in \fig{\ref{fig:multi-object-discrimination}} (b), given a proto-map $\mathbf{m}^{i}$ and the condition code $\mathbf{c}^{i}$, the mask calibration is given by:
\begin{align}
        \mathbf{m}^{i}_{out} & = \theta_{dec}( \mathbf{m}^{i}_{in} + \mathbf{m}^{i}_{in} \otimes \mathbf{c}^{i} ).
\end{align}        

We cascade $N_{L}$ combinations of discriminative object calibrators and conditional decoders to progressively refine and up-sample the proto-map $\mathbf{m}^i$ for each object $i$. Besides, the proto-maps are concatenated with the low-level features of the current frame from the backbone in the second-to-last conditional decoder, introducing more fine-grained pixel-level image cues for mask calibration. As is shown in \fig{\ref{fig:multi-object-discrimination}} (c), we can find the proto-map evolves to be more discriminative, \ie, cues of other (even similar) objects are substantially suppressed, thus producing a more accurate object mask for the target object.

\renewcommand{\arraystretch}{0.68}
\begin{table*}[t]
\centering
	{
	\begin{tabular}{l|ccc|ccccc|ccccc}
	\toprule
	\multirow{3}{*}{Methods} && & & \multicolumn{5}{c|}{YouTube-VOS 2018 Validation}  & \multicolumn{5}{c}{YouTube-VOS 2019 Validation}                                    \\
	\cmidrule(l){2-14} &$AF$ &$MF$ & $EXD$
	& $\mathcal{J}\&\mathcal{F}$ & $\mathcal{J}_{s}$ & $\mathcal{J}_{u}$ & $\mathcal{F}_{s}$ & $\mathcal{F}_{u}$ & $\mathcal{J}\&\mathcal{F}$ & $\mathcal{J}_{s}$ & $\mathcal{J}_{u}$ & $\mathcal{F}_{s}$ & $\mathcal{F}_{u}$\\
	\midrule 
	PReM {\cite{luiten2018premvos}} & & &&   66.9 & 71.4 & 56.5 & 75.9  & 63.7 & -     & -& -     & -  & -   \\
	CFBI {\cite{yang2020collaborative}}&&&&    81.4 & 81.1 & 75.3 & 85.8  & 83.4  &81.0 & 80.6 & 75.2 & 85.1  & 83.0\\
	CFBI+ {\cite{yang2021collaborative}}&&&&   82.8 & 81.8 &77.1&86.6&85.6 & 82.6& 81.7 &86.2& 77.1& 85.2    \\
    {AOT-B} {\cite{yang2021associating}}&&&\checkmark&    83.2 & {82.6} &{77.3}&\textbf{87.4}&85.6 & 83.3 & \textbf{82.5} &77.8&\textbf{87.0}&86.0     \\	
	\textbf{Ours-Base} & &      &  & {{{\textbf{83.6}}}} & \textcolor{black}{{\textbf{82.6}}} & \textcolor{black}{{\textbf{78.3}}} & \textcolor{black}{{87.2}}  & \textcolor{black}{{\textbf{86.3}}} &   \textcolor{black}{{\textbf{83.7}}} & \textcolor{black}{{82.3}} & \textcolor{black}{{\textbf{{79.0}}}} & \textcolor{black}{{86.6}}  & \textcolor{black}{{\textbf{86.9}}}\\
    \midrule \midrule 
	STM {\cite{oh2019video}}&&\checkmark&\checkmark&      79.4 & 79.7 & 72.8 & 84.2  & 80.9 & -     & -& -     & -  & -   \\
	LCM {\cite{hu2021learning}}&&\checkmark&\checkmark&  82.0&82.2&75.7& {86.7}&83.4 & -     & -& -     & -    & -  \\
	MiVOS+km {\cite{cheng2021modular}}&&\checkmark&\checkmark&  82.6 & 81.1 &77.7&85.6&86.2 &  82.8 & 81.6 &77.7&85.8&85.9       \\
    DMN-AOA \cite{9710441}& &\checkmark&\checkmark& 82.7 & 82.6 & 76.7 & 87.0 & 84.8 & - &- &- &- &-    \\	
	STCN {\cite{cheng2021rethinking}}&\checkmark&\checkmark&\checkmark&    83.0 & 81.9 &{77.9}&86.5&85.7 & 82.7 & 81.1 &78.2&85.4&85.9     \\	
	JOINT {\cite{mao2021joint}}&&\checkmark&&    83.1 & 81.5 &{78.7}&85.9&86.5 & 82.8 & 80.8 &79.0&84.8&86.6     \\

    {AOT-L} {\cite{yang2021associating}}&&\checkmark&\checkmark&    83.7 & 82.5 &{77.9}&\textbf{87.5}&86.7 & 83.6 & 82.2 &78.3&86.9&86.9     \\
	\textbf{Ours-MF} &&\checkmark &      &   {{{\textbf{84.0}}}} & \textcolor{black}{{\textbf{82.7}}} & \textcolor{black}{{\textbf{78.8}}} & \textcolor{black}{{{87.4}}}  & \textcolor{black}{{\textbf{87.1}}} &   \textcolor{black}{{\textbf{84.1}}} & \textcolor{black}{{\textbf{82.7}}} & \textcolor{black}{{\textbf{79.4}}} & \textcolor{black}{{\textbf{86.9}}}  & \textcolor{black}{{\textbf{87.2}}}\\
	\midrule \midrule 
	CFBI${}^{MS}$ {\cite{yang2020collaborative}}& &&&   82.7 & 82.2  & 76.9 & 86.8  & 85.0  &      82.4 & 81.8 & 76.9 & 86.1  & 84.8   \\	
	CFBI+${}^{MS}$\cite{yang2021collaborative} &&   &    & 83.3 & 82.8  & 77.3 & 87.5 & 85.7 &-&-&-&- &- \\
		\textbf{Ours-Base}${}^{MS}$ &&&&\textbf{84.4} & \textbf{83.2}&\textbf{79.3}&\textbf{87.8} &\textbf{87.3}  & \textbf{\textbf{84.4}} & \textbf{\textbf{82.7}}  & \textbf{80.0} &\textbf{\textbf{87.1}}  &    \textbf{\textbf{87.8}}  \\
	\bottomrule
	\end{tabular}
	}
	\caption{Quantitative comparison on YouTube-VOS \cite{xu2018youtube}. $AF$ denotes using All-Frames (30FPS) videos instead of default (6FPS) videos. $MF$ denotes multiple historical frames are leveraged as guidance for current frame, otherwise only using the first and the previous frame. $EXD$ denotes using external (static image transformation) data for training. $MS$ denotes using multi-scale and flip testing in evaluation.}
	\label{table:ytb}
\end{table*}

\subsection{Network Details}

\noindent\textbf{Correlation-based Proto-map Generation.}
Given the overall adaptive object proxy map $\mathbf{p}^i = [...; \mathbf{p}_{r}^{i,K}; ...]$ for each object $i$ as discussed in Sec.{\ref{sec:adaptove_proxy_representation}}, 
we first calculate the similarity maps between the queries of the current feature map $\mathbf{f}_{t}$ and each element $\mathbf{p}_{r}^{i,K}$ separately.
Then, the proto-map $\mathbf{m}^i$ is generated by translating the concatenation of all the similarity maps and current feature map $\mathbf{f}_{t}$. Formally,
\begin{small}
\begin{equation}
    \mathbf{m}^i = \theta_{ens} ([\varphi_s ([...; Sim(\mathbf{f}_t,  \mathbf{p}_{r}^{i,K}; ...]),\mathbf{f}_t]),  \quad r \in \mathcal{S}_{ref}, K \in \mathcal{L}_{clu}
\end{equation}
\end{small}
where $\theta_{ens}(\cdot)$ consists of the first two stages of bottleneck blocks in the ensembler of \cite{yang2020collaborative}, $\varphi_s$ is a $1\times1$ convolution layer to project the similarity maps in the channel dimension, and $Sim(\cdot,\cdot)$ is a L2-norm-based similarity function.

		
\noindent\textbf{Conditioning Layer.}
\label{sec:method/relational_modulation_network/conditioning_layer} Conditioning Layer ($CL$) is a block in the discriminative object calibrator to calculate the condition code $\mathbf{c}$ from a feature map $\mathbf{z}_{in}$, taking the form:
\begin{small}
\begin{align}
	 CL(\mathbf{z}_{in}) =  MLP (GAP(\mathbf{z}_{in} \odot \pi_{\beta}( \varphi (\mathbf{z}_{in}) )),~
	\pi_{\beta}(x) = 
	\begin{cases}
		x & x \ge \beta \\
		0 & x < \beta
	\end{cases}
\end{align}
\end{small}
where $\varphi (\cdot)$ represents a $1\times 1$ convolution layer with ReLU activation to project the input $\mathbf{z}_{in}$ to a confidence map, $GAP$ denotes global average pooling. 
We incorporate a confidence gate $\pi_{\beta}$ in the conditioning layer to filter out unreliable cues in the input feature map, where $\beta$ is chosen as a percentile value in $\varphi(\mathbf{z}_{in})$.

\newcommand{\tabincell}[2]{\begin{tabular}{@{}#1@{}}#2\end{tabular}}
\begin{table*}
	\centering
	\setlength{\tabcolsep}{0.6mm}
	\begin{small}
	\resizebox{1\textwidth}{!}
	{
    \begin{tabular}{c|c||ccccccc|c||cccccccccc}
    \toprule

             \multicolumn{2}{c||}{Reference frames}  &\multicolumn{8}{c||}{$1$\&$(t-1)$ frames}   & \multicolumn{9}{c}{Multiple frames}     \\ \cmidrule{1-19}
           \multicolumn{2}{c||}{\multirow{2}*{Method}}
          &OnAVOS${}^{*}$&RGMP${}^{*}$&FEEL&PReM${}^{*}$& CFBI&AOT-B&\textbf{Ours}&\textbf{Ours${}^{FR}$}&STM & SST${}^{*}$ & MiVOS & RMN & LCM  & JOINT& HMMN &STCN &\textbf{Ours}\\
          \multicolumn{1}{c}{}& \multicolumn{1}{c||}{}&\cite{voigtlaender2017online}&\cite{oh2018fast}&\cite{voigtlaender2019feelvos}&\cite{luiten2018premvos}& \cite{yang2020collaborative}&\cite{yang2021associating} &\textbf{-Base}&\textbf{-Base} &\cite{oh2019video} & \cite{duke2021sstvos} & \cite{cheng2021modular} & \cite{RMN} & \cite{hu2021learning}  &\cite{mao2021joint}&\cite{seong2021hierarchical}&\cite{cheng2021rethinking} &\textbf{-MF} \\
          	\midrule
    \multirow{3}*{\tabincell{c}{Davis16\\Valid }}&    $\mathcal{J}\&\mathcal{F}$ & 85.0&81.8&81.7&86.8& 89.4 &89.9 & \textbf{90.7}&\textbf{91.2}&89.3   & - & {91.0} & 88.8& 90.7 &{-}&90.8&{91.6} & \textbf{91.6} \\ 
    &    $\mathcal{J}$  &85.7& 81.5&81.1&84.9&88.3&\textbf{88.8}& 87.1&\textbf{88.0}& 88.7    & - & {89.7} & 88.9 &89.9 & - &89.6&\textbf{90.8} & 88.5 \\ 
    &    $\mathcal{F}$  &84.2&82.0&82.2&88.6& 90.5&90.9&\textbf{94.2}&\textbf{94.4}&89.9    & - & 92.1         & 88.7  &91.4& 93.9 &92.0&92.5 & \textbf{94.7}\\ \midrule \midrule 
        \multirow{3}*{\tabincell{c}{Davis17\\ Valid}}&$\mathcal{J}\&\mathcal{F}$  &65.4&66.7&71.5&77.8& 81.9&82.1& \textbf{{83.1}}&\textbf{84.0}&81.8   & 82.5 & 83.3 & 83.5 & 83.5 & 83.5 &84.7&\textbf{85.4} &83.8  \\ 
        &$\mathcal{J}$  &61.6&64.8&69.1&73.9& 79.1&79.4 & \textbf{{80.5}}&\textbf{{81.0}}&79.2    & 79.9 & 80.6 & 81.0 & 80.5 & 80.8&81.9&\textbf{82.2} & 81.7  \\ 
        &$\mathcal{F}$  &69.1&68.6&74.0&{81.8} &84.6& 84.8    &\textbf{85.7}&\textbf{{86.9}}&84.3   & 85.1 & 85.1 & 86.0 & {86.5} & 86.2 &87.5&\textbf{88.6} & 85.9\\ 
        \midrule \midrule 
        \multirow{3}*{\tabincell{c}{Davis17\\ Test-dev}}&$\mathcal{J}\&\mathcal{F}$ &52.8&52.9&57.8&71.6& 74.8 &75.5& \textbf{{76.5}}&\textbf{{77.5}}&72.3    & -  & 76.5 & 75.0  & 78.1 & - &78.6&76.1& \textbf{79.3} \\ 
        &$\mathcal{J}$ &49.9&51.3&55.1&67.5& 71.1 &71.8& \textbf{{72.4}}&\textbf{{73.6}} &69.3     & - & 72.7  & 71.9 & 74.4 & - &{74.7}&72.7& \textbf{74.7} \\ 
        &$\mathcal{F}$  &55.7&54.4&60.4&75.7& 78.5 &79.1& \textbf{{80.6}}&\textbf{{81.3}} &75.2   & - & 80.2  & 78.1 & 81.8 & - &82.5&79.6& \textbf{83.9}\\        
        \bottomrule
    \end{tabular}
    }\end{small}
\caption{Quantitative comparisons on DAVIS16\cite{perazzi2016benchmark} and DAVIS17\cite{pont20172017}. $FR$ denotes full-resolution testing, otherwise methods are tested on $480p$. $*$ denotes training with DAVIS only, otherwise with both DAVIS and YouTube-VOS.}
    \label{table:dv}
\end{table*}

\section{Experiment}\label{sec:Experiment}
To evaluate the performance and robustness of our model, we conduct experiments on both standard VOS benchmarks and our constructed perturbed benchmark.
\subsection{Standard Benchmark}
\noindent\textbf{Datasets.}
We evaluate our method on two widely-used multi-object VOS benchmarks, 
    \textit{i.e.},
    YouTube-VOS~\cite{xu2018youtube} and DAVIS17~\cite{pont20172017}. The unseen object categories make YouTube-VOS a good benchmark to 
    measure the generalization ability of various methods.
Besides,
    the comparison between our method and other methods on a single-object VOS benchmark DAVIS16~\cite{perazzi2016benchmark} will also be included.

\noindent\textbf{Metrics.} 
We adopt the evaluation metrics from DAVIS~\cite{perazzi2016benchmark},
    \textit{i.e.},
    the region accuracy $\mathcal{J}$ and boundary accuracy $\mathcal{F}$. $\mathcal{J}$ measures the intersection-over-union (IoU) between the predicted masks and the ground-truth masks, 
    and $\mathcal{F}$ measures the accuracy of masks on the boundaries via bipartite matching between the boundary pixels. 
For both metrics,  we will report the performance on seen and unseen categories as $\mathcal{J}_{s}$, $\mathcal{J}_{u}$ and $\mathcal{F}_{s}$, $\mathcal{F}_{u}$ respectively.

\subsection{Pilot Robustness Benchmark}
\label{sec:robustness_benchmark}
\noindent\textbf{Perturbed datasets.}
For a type of perturbation ${\epsilon} \sim E$, we can generate a perturbed dataset $\mathcal{D}_{{\epsilon}}={\epsilon}(\mathcal{D})$. Concretely, we established a pilot validation benchmark, namely \textit{YouTube-VOS-P}, to evaluate VOS robustness against image perturbations based on the clean YouTube-VOS-2019 \cite{xu2018youtube} validation set, which is the largest multi-object VOS dataset. In \textit{YouTube-VOS-P}, we apply 6 types of perturbations on clean data for perturbed dataset construction. The perturbation set is constructed with noises \cite{deledalle2012compare} and blurring \cite{liu2020estimating}, which widely exist in real-world videos. Specifically, perturbation types include \textit{Gaussian blur with $7\times7$ and $9\times9$ kernel}, \textit{salt and pepper noise} with 1$k$ or 5$k$ points, and \textit{Gaussian noise with mean of 0 and standard deviation of 10 or 30}. All perturbations are implemented with OpenCV \cite{bradski2008learning}.

\noindent\textbf{Robustness metrics.} 
Following robustness evaluation metrics commonly used in other tasks \cite{hendrycks2019benchmarking,kamann2020benchmarking,laugros2019adversarial,tramer2019adversarial,jin2019bert}, we put forward two metrics for the evaluation of robustness against perturbation for a VOS model as follows.

\noindent\textbf{After-perturbation accuracy ($\mathcal{Q}_p$).} After-perturbation accuracy $\mathcal{Q}_p$ represents the averaged remaining overall performance after perturbation. Specifically, given all the perturbation operation ${\epsilon} \sim E$, the after-perturbation accuracy is defined as
\begin{equation}
  \mathcal{Q}_p = \frac{1}{|E|}\sum_{\epsilon \sim E}\mathcal{Q}_{{\epsilon}}.   
\end{equation}
\noindent\textbf{Perturbation robustness ($\mathcal{R}_p$).} Given the performance $\mathcal{Q}_c$ on the original clean dataset and the after-perturbation accuracy ($\mathcal{Q}_p$) on the perturbed dataset, we can approximate the overall robustness to perturbation $\mathcal{R}_p$ for a VOS model with the average overall performance drop, which can be formulated as 
\begin{equation}
\mathcal{R}_p = \mathcal{Q}_c-\mathcal{Q}_p. 
\end{equation}
Here, smaller $\mathcal{R}_p$ indicates better robustness for a VOS model. 

\subsection{Implementation Details.}
We use SGD \cite{bottou2010large} with momentum 0.9 as the optimizer and use cross-entropy loss following our baseline setting~\cite{yang2020collaborative}.
For all the experiments, we set the batch size to 8.
For training on YouTube-VOS, we do not use any external data.
The total training iterations is 400$k$. We set
the learning rate as 0.02 for the first half (200$k$) and 0.01 for the rest.
We use the DeepLabv3+~\cite{chen2018encoder} architecture with ResNet-101~\cite{he2016deep} as the backbone of our model. For multi-scale inference, 
we apply a scale set of [1.0, 1.15, 1.3, 1.5] as previous works \cite{yang2021collaborative,yang2020collaborative}.
For training on DAVIS~\cite{perazzi2016benchmark}, 
    we finetune the model pre-trained on YouTube-VOS for 40$k$ iterations with learning rate of 0.1 and we use both DAVIS and YouTube-VOS datasets with a sampling ratio of 2:1. An NVIDIA Linux workstation (GPU: 8$\times$ Tesla V100) is used for our experiments. Our codebase is built on PyTorch 1.8.0~\cite{paszke2019pytorch} and 
    reuses some components implemented in~\cite{yang2020collaborative}. 
We set multi-run clustering list $\mathcal{L}_{clu}$ = [1, 16, $|H\times W|$]. The parameter $\beta$ used in the conditioning layer as discussed in 
    Sec.~\ref{sec:method/relational_modulation_network/conditioning_layer} is $0.3$ through grid search. The number of cascaded mask decoding blocks ${N_L}$ is set to 6. 

 We set two model variants for a fair comparison with previous methods: (1) \textbf{Ours-Base} is the default setting with the first and previous frame as reference, \textit{i.e.}, $\mathcal{S}_{ref}=\{1,t-1\}$ as mentioned in Section \ref{sec:method/overview}; (2) \textbf{Ours-MF} uses multiple historical frames following ~\cite{oh2019video,yang2021associating}, \textit{i.e.}, $\mathcal{S}_{ref}=\{1, 1+\delta, 1+2\delta, 1+3\delta, ...\}, \delta=5$.

\subsection{Result on Standard VOS Benchmark}
\textbf{Quantitative comparison.}
The comparison between our method and other state-of-the-art methods
    on YouTube-VOS 2018 and YouTube-VOS 2019 validation set
        is shown in Table~\ref{table:ytb}, which shows
    our model outperforms all the previous SOTA methods even without using any external data.
Our method stands out under various evaluation metrics, especially on unseen categories, 
    which demonstrates the great generalization ability of our method. The comparison on DAVIS is provided in Table~\ref{table:dv}. Our model also achieves the best performance on the challenging DAVIS17 test-dev split.
    
\begin{figure*}[t!]
	\centering
	\includegraphics[width=\textwidth]{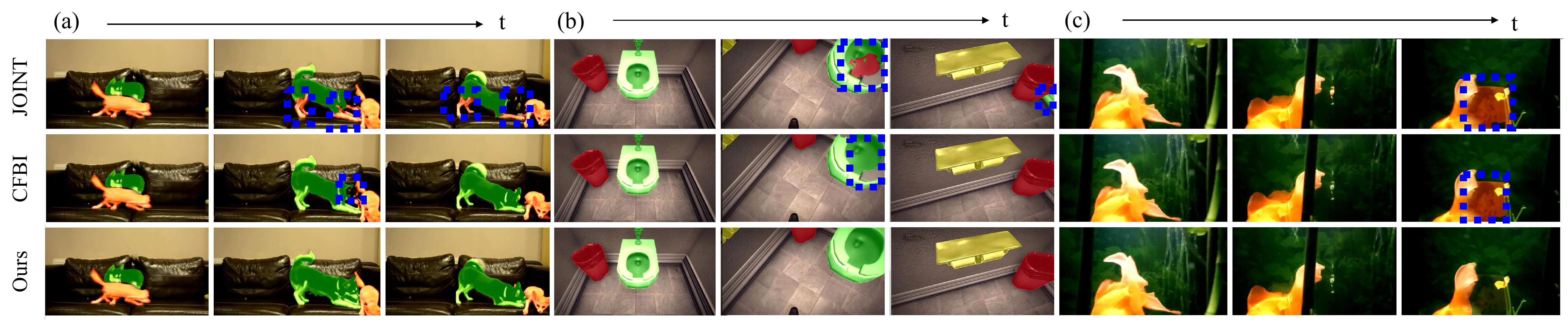}
\caption{Qualitative comparison to competitive methods, JOINT and CFBI on YouTube-VOS 19 validation set. With the proposed adaptive proxy representation and object mask calibration for mask decoding, our model can tackle cases such as (a) object occlusion, (b) large camera rotation, and (c) fast motion better. Error regions are highlighted with blue bounding boxes.}
	\label{fig:qualitative_comparison}
\end{figure*}

\begin{figure}[t]
\centering
\includegraphics[width=0.42\textwidth]{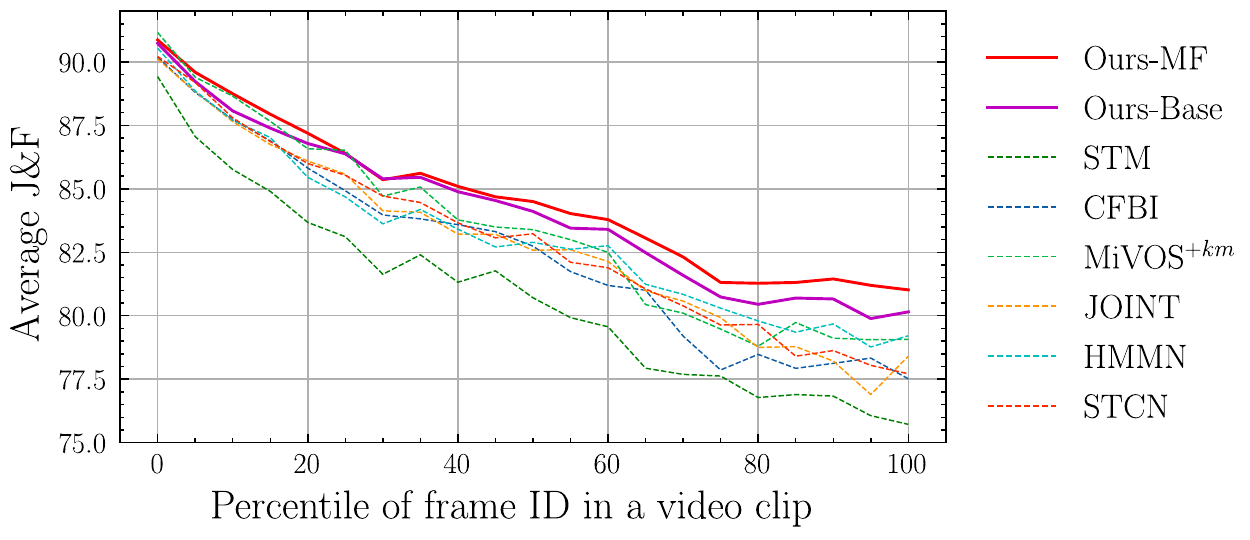}
\caption{Comparison of average overall performance ($\mathcal{J}\&\mathcal{F}$) on YouTube-VOS19 over time. 0(\%) and 100(\%) represents the beginning and the end of a video clip respectively. Performance on all the videos are normalized to the same length. Ours has the least performance decay.} \label{fig:YTB_decay}                   
\end{figure}

\noindent\textbf{Qualitative comparison.}
Fig.~\ref{fig:qualitative_comparison} shows the qualitative comparison between previous SOTA methods and our model (Ours-Base) under various hard cases, and our method performs well for all these cases.
The accuracy changes of different methods over time on YouTube-VOS 2019 validation set are illustrated in Fig~\ref{fig:YTB_decay}. Our method has the least performance decay, demonstrating better robustness to suppress error propagation.


\renewcommand{\arraystretch}{1}
\begin{table}[t]
	\resizebox{0.48\textwidth}{!}
	{
\begin{tabular}{@{}l|ccc|ccc@{}}
\toprule%
         Reference frames   & \multicolumn{3}{c|}{$1$\&$(t-1)$ frames}   & \multicolumn{3}{c}{Multiple frames}     \\ \midrule
             
         \multirow{2}{1in}{Methods}    & \multicolumn{1}{c}{\textbf{Ours}} & {CFBI} & {AOT-B}    & {\textbf{Ours}}  & {AOT-L}    & {STCN}     \\
             & \multicolumn{1}{c}{\textbf{-Base}} & {\cite{yang2020collaborative}} & {\cite{yang2021associating}}    & {\textbf{-MF}}  & {\cite{yang2021associating}}    & {\cite{cheng2021rethinking}}     \\
             \cmidrule{1-7}
Clean accuracy $\mathcal{Q}_c$  $\uparrow$     & \textbf{83.7}      & 81.0                   & 83.3                     & \textbf{84.1}               & 83.6                     & 82.7                       \\ \midrule  \midrule
 +Gaussian noise ($\sigma=10$)     &    \textbf{83.1}  & 80.5       & 82.7                       &     \textbf{83.4}             &   82.8                         &      80.8                  \\
+Gaussian noise ($\sigma=30$) & \textbf{80.8}  &    76.6   
           &   77.0     & \textbf{81.1}                       &   77.6                       &   78.6                                               \\ 
+Salt\&pepper noise (1$k$)   &   \textbf{83.3}               & 80.0       & 83.2                  &          \textbf{83.8}                   &   83.4                         &    80.1                      \\
+Salt\&pepper noise (5$k$)  & \textbf{82.2}           & 79.1                          &  82.0                          &   \textbf{82.5}                    & 81.4                     &  78.0                          \\
+Gaussian blur ($7\times7$)   & {82.7} 
             & 80.4      &  \textbf{82.8}  & \textbf{83.0}     & 82.8                        & 80.9                          \\
+Gaussian blur ($9\times9$)    & \textbf{82.0}    & 79.9                           &  81.7                     &   \textbf{82.4}                          &  82.2                          & 79.9                        \\\midrule
After-perturbation accuracy $\mathcal{Q}_p$ $\uparrow$   & \textbf{82.3}  	
             & 79.4     & 81.6     & \textbf{82.7}    & 81.7                       & 79.7                       \\ \midrule\midrule
Perturbation robustness $\mathcal{R}_p$ $\downarrow$  & \textbf{1.4}  	
             &  1.6    & 1.7     & \textbf{1.4}    & 1.9                    & 3.0                       \\ 
 \bottomrule
\end{tabular}}
    	\caption{Pilot study of perturbation robustness for VOS models on the perturbed dataset {YouTube-VOS-P}. We also list the official results on {YouTube-VOS} (Clean) for reference. Results are reported with $\mathcal{J}\&\mathcal{F}$ .}
    	\label{tab:robustness}
\end{table}
\subsection{Analysis on Robustness to Perturbation.}

\noindent In our pilot study on VOS model robustness, the proposed perturbed VOS validation dataset, \textit{YouTube-VOS-P}, is used for robustness evaluation, as introduced in Sec.\ref{sec:robustness_benchmark}. Note that all the models to be evaluated are trained with clean datasets. The results under random perturbations for AOT-B \cite{yang2021associating}, AOT-L \cite{yang2021associating}, and STCN \cite{cheng2021rethinking} (without using external BL30K dataset\cite{cheng2021modular}) are evaluated with their official model checkpoints while CFBI \cite{yang2020collaborative} is retrained by us and the retrained result on the original clean dataset matches the performance reported in their paper. Experiments are averaged for 3 runs for validity. The results are summarized in Table~\ref{tab:robustness} and we provide the insights and discussions as follows.

\noindent\textbf{\textit{Are current VOS models robust to image perturbations?}} No. We notice a clear performance drop for all the VOS models investigated. Noticeably, these models are vulnerable to attack even when the input is injected with simple random Gaussian noise. Meanwhile, we can notice that the after-perturbation performance will drop largely as the severity of perturbations increases.

\noindent\textbf{\textit{Does a model with higher performance on clean benchmarks guarantee better robustness against perturbations?}} Not sure. When making a comparison across various models, the correlation of the average overall performance ($\mathcal{Q}_c$) on the original clean YouTube-VOS dataset and the after-perturbation accuracy ($\mathcal{Q}_p$) is not clear. When making comparisons within a certain model, higher performance for one model also may not ensure stronger robustness because prediction errors and noises may be propagated during correlation calculation for VOS. 

\begin{figure}[t]
\centering
\includegraphics[width=0.48\textwidth]{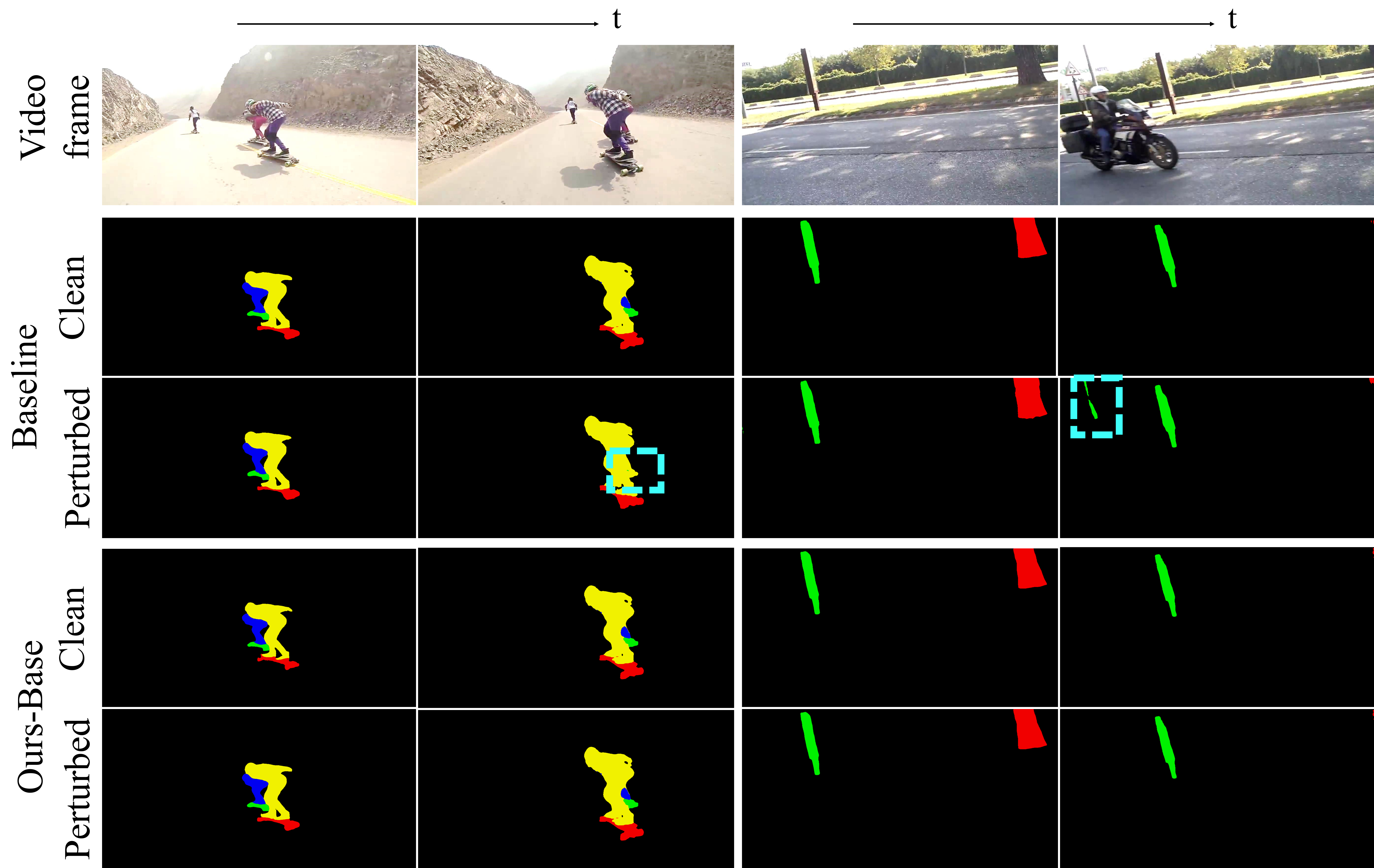}

\caption{Qualitative comparison between our model (Ours-Base) and the baseline model \cite{yang2020collaborative} on robustness against perturbations. Two models performs well on original clean videos, but our model outperforms the baseline on perturbed videos with Gaussian Noise.}
\label{fig:qualitative_robust}                   
\end{figure}

\noindent\textbf{\textit{Can the proposed VOS framework help improve perturbation robustness?}} Yes! Our VOS framework with adaptive proxy aggregation and multi-object discrepancy discrimination mechanism stands out in the perturbed dataset setting, achieving higher after-perturbation accuracy $\mathcal{Q}_p$ (82.3 \textit{V.S.} 79.4 in $\mathcal{J}\&\mathcal{F}$) and better perturbation robustness $\mathcal{R}_p$ (1.4 \textit{V.S.} 1.6 in $\mathcal{J}\&\mathcal{F}$) than CFBI which is our baseline model. Though our baseline is strong enough to well handle some cases with multiple similar objects on the original clean datasets, the prediction of our model is more stable and consistent under perturbations, as is shown in Fig.~\ref{fig:qualitative_robust}.

\subsection{Ablation Study}
\label{sec:ablation_study}
We make a thorough analysis of various components used in our method.
All the ablation studies 
    are based on our default setting (\textbf{Ours-Base}) as mentioned in Section \ref{sec:method/overview} and conducted on the YouTube-VOS 2019 validation set.


\renewcommand{\arraystretch}{1}
\begin{table}[t]

	\centering
	\resizebox{0.48\textwidth}{!}
	{
    \begin{tabular}{@{}l|ccccc|c|c@{}}\toprule
    Model          & \multirow{1}*{\tabincell{c}{$\mathcal{J}\&\mathcal{F}$}}  & \multirow{1}*{\tabincell{c}{$\mathcal{J}_s$}} & \multirow{1}*{\tabincell{c}{$\mathcal{J}_u$}} & \multirow{1}*{\tabincell{c}{$\mathcal{F}_s$}} & \multirow{1}*{\tabincell{c}{$\mathcal{F}_u$}}&FPS&Param \\ 

    \midrule
Baseline\cite{yang2020collaborative}  & 81.0 & 80.6    & 75.2      &
	{85.1}    & 83.0 &3.4&66.1 \\ \midrule
GSPR & 82.4 & 81.6 & 77.4 & 85.8 & 85.0 &3.3&66.1 \\
AOP ($K$=8) &  82.8 & 82.1  & 77.6  & 86.5  & 85.0 &3.2&66.1 \\
AOP  ($K$=16, ours)   & 82.9 & 82.1    & 77.8      &
	{86.4}    & 85.2 &3.2&66.1 \\
AOP ($K$=32) & 82.8 & 82.2   &77.6   & 86.4  & 85.1 &3.2&66.1  \\
AOP ($K$=256) & 82.2  & 81.6  & 77.4  & 86.0  &  83.8 &3.0&66.1 \\   \midrule 
	DOC ($\beta=0$) & 83.0 & 82.2    & 77.9      & 86.5    & 85.4  &3.3&66.7    \\ 
         DOC ($\beta=0.3$, ours)   & 83.4   & 82.1 & 78.4 & 86.5 & 86.3 &3.3&66.7 \\\midrule
  
    \textbf{Ours-Base}   & \textbf{83.7} & \textbf{82.3} & \textbf{79.0} & \textbf{86.6} & \textbf{86.9} &3.2&66.7 \\ 
  
     \bottomrule
    \end{tabular}
    }
    \caption{Ablation study of the proposed adaptive object proxy (AOP) representation and discriminative object calibration (DOC). Here AOP and DOC denote models using AOP or DOC only. Grid-sampling-based proxy representation (GSPR) is for comparison with AOP. The inference time is reported in \textit{multi-object FPS} as previous works \cite{yang2020collaborative,yang2021associating} and measured on a single V100 NVIDIA GPU with $batchsize=1$. The number of model parameters is reported in MB.}
    \label{table:component_ablation}

\end{table}

\noindent\textbf{Component effectiveness study.}
Table~\ref{table:component_ablation} 
    demonstrates the effectiveness of the proposed adaptive object proxy (AOP) representation and 
        discriminative object calibrator (DOC) for mask decoding. 
The first row shows the result of our baseline method ~\cite{yang2020collaborative}  
    which adopts pixel-level correlation and uses modulation during mask decoding. 

\begin{figure}[t]
	\centering
	\includegraphics[width=0.48\textwidth]{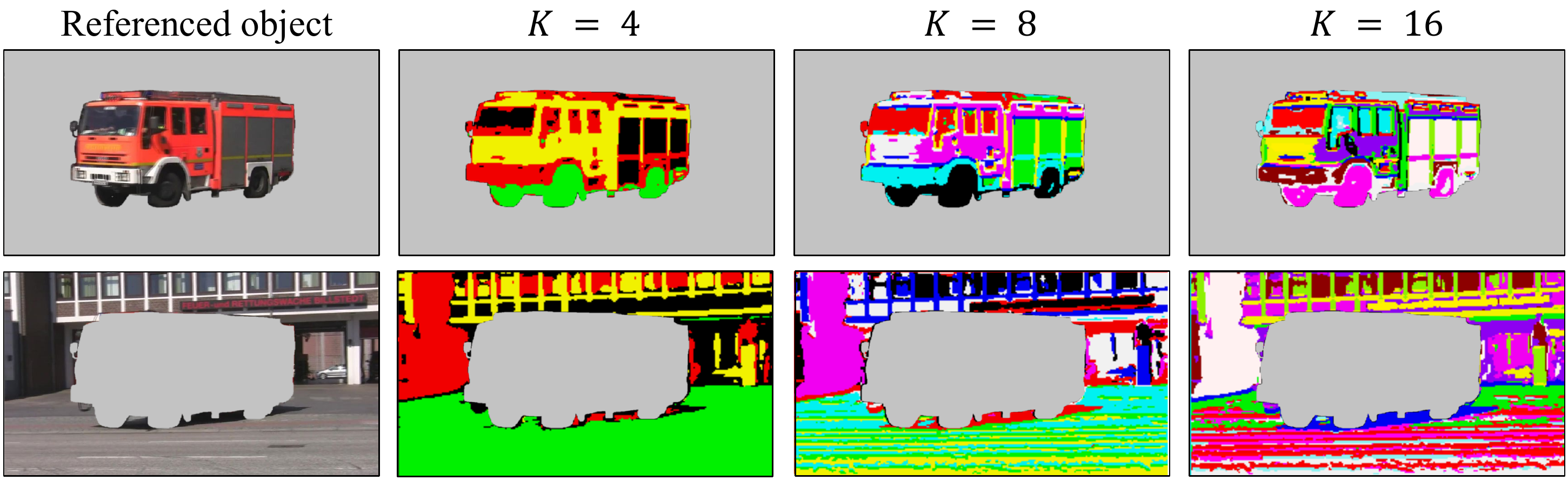}
\caption{Visualization of clustering segments of different proxy number $K$ for the construction of adaptive proxy representation. Semantically similar pixels in the foreground or background are grouped into semantic parts, such as wheels of the car and windows in the background.}
	\label{fig:clustering_examples}
\end{figure}

\noindent\textbf{Ablation on adaptive object proxy (AOP) representation.}
Compared to the commonly-used pixel-level object representation used in the baseline model, our adaptive proxy representation formed with help of clustering-based aggregation can significantly improve the overall performance, especially for unseen categories (+2.6\% in $\mathcal{J}_u$ and +2.2\% in $\mathcal{F}_u$). Meanwhile, our adaptive proxy representation also performs better than simply aggregating the features grid-to-grid to form grid-sampled proxies (GSPR). Fig.~\ref{fig:clustering_examples} illustrates the segments constructed from K-Means clustering algorithm, as we can observe, each segment corresponds to a meaningful semantic part of an object, which can be aggregated to robust proxies for further correlation calculation across frames.

\noindent\textbf{Ablation on discriminative object calibrator (DOC) for mask decoding.}The multi-object discrimination mechanism help boost the performance from 82.9\% to 83.7\% given the model with adaptive proxy representation. Meanwhile, the confidence gate of the conditioning layer in DOC can also help improve the overall performance (+0.4\% in $\mathcal{J}\&\mathcal{F}$) compared to the setting when disabling the confidence gate ($\beta$=0), which is owing to its de-noising mechanism to filter unreliable cues.

\noindent\textbf{Complexity analysis.} As is shown in the running time and model size ablation study in Table \ref{table:component_ablation}, compared to our baseline, our model achieves much better overall performance (83.7\% \textit{V.S.} 81.0\% in $\mathcal{J}\&\mathcal{F}$) with a negligible cost of inference speed (-0.2 multi-object FPS) and model parameters (+0.6 MB).

\begin{figure}[t!]
	\centering
	\includegraphics[width=0.48\textwidth]{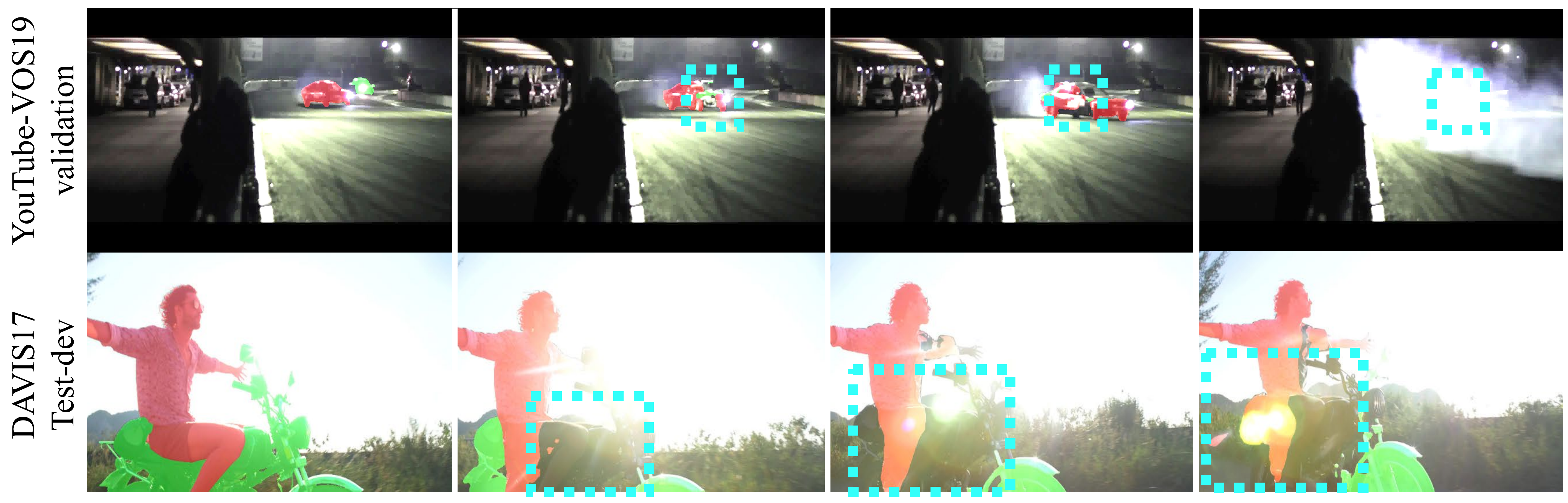}
\caption{Limitation. We show two failure cases with strong natural perturbations, including the smoke (top) and the blur caused by a strong halo (bottom), of our model.}
	\label{fig:failure_case}
\end{figure}

\section{Conclusion}
\label{sec:Conclusion}
Our pilot study on model robustness to perturbations for VOS reveals the fragility of current advanced methods. Towards robust VOS, we propose an end-to-end network with two tightly coupled modules to generate adaptive proxy representation and perform object mask calibration with multi-object discrimination considered. 
The two modules are demonstrated to help achieve a significant gain on standard VOS benchmarks and better robustness against perturbations compared to the baseline method.

\noindent\textbf{Limitation and discussion.}
Though the performance of our adaptive proxy aggregation with non-parametric $K$-Means clustering is not sensitive to the number of clusters, we consider it is better to design a learning-based way to generate adaptive parts. 
 Moreover, as perturbations in natural videos are more diverse, such as the smoke and the severe blur shown in Fig.~\ref{fig:failure_case}, perturbation types can be extended to more diverse natural corruptions and adversarial attacks. For the robustness enhancement of VOS models, the data organization part can be further explored.

\clearpage
\bibliographystyle{ACM-Reference-Format}
\bibliography{main}

\clearpage

\setcounter{table}{0}  
\setcounter{figure}{0}  
\setcounter{section}{0}  
\renewcommand{\thetable}{\Alph{table}}
\renewcommand{\thefigure}{\Alph{figure}}
\renewcommand{\thesection}{\Alph{section}}

\section{More Qualitative Results}

More qualitative comparisons to the state-of-the-art models \cite{cheng2021rethinking,yang2020collaborative,oh2019video} and ablation studies on our two main components, our adaptive object proxy representation and discriminative object calibration, are presented in the video.\footnote{The video demo is available at https://youtu.be/3F6n7tcwWkA}
\subsection{Qualitative Results on Video Clips with An Extremely Large Number of Objects}
\begin{figure*}[t]
	\centering
	\includegraphics[width=0.94\textwidth]{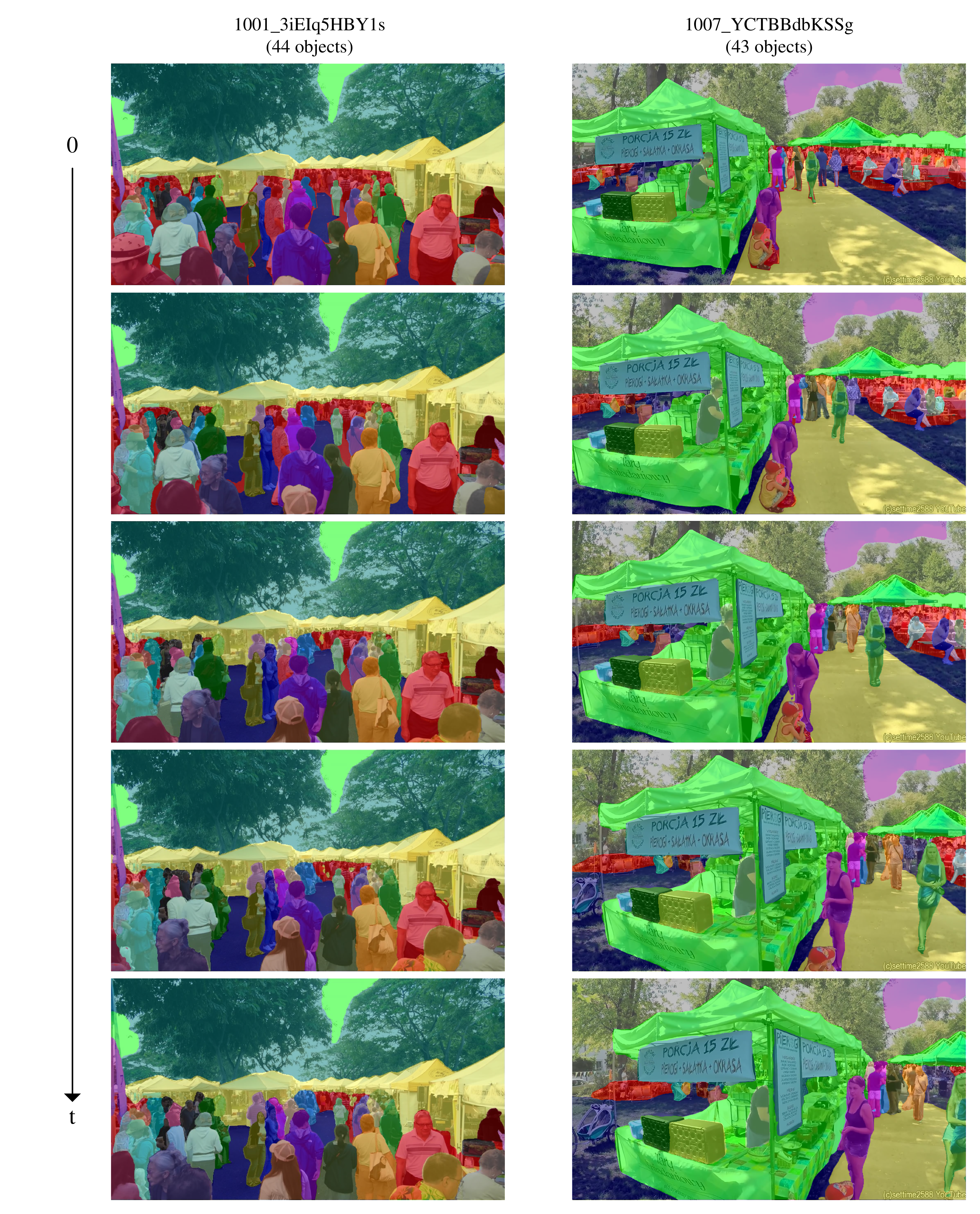}
\caption{We show two cases with an extremely large number of target objects to be segmented, including \textit{1001\_3iEIq5HBY1s} with 44 objects (left) and \textit{1007\_YCTBBdbKSSg} with 43 objects (right), from a video scene parsing dataset, VSPW \cite{miao2021vspw}. Even if our model (Ours-Base) only leverages the guidance from the first and the previous frame, it can handle such cases with an extremely large number of objects well. Please zoom in on the figure to view it better.}
	\label{fig:vspw}
\end{figure*}
In Fig.~\ref{fig:vspw}, we use our model (Ours-Base) to propagate the masks in two cases where more than 40 objects co-exist. We leverage the video clips from a video scene parsing dataset, VSPW\cite{miao2021vspw}, and use the mask of the first frame as the reference for mask propagation. Even though we only segment each video frame with the guidance from the first frame and the previous frame, we can segment most objects in such a crowded scene well. Notice that we infer the video clip in $480p$ and the given mask in the first frame is not accurate enough especially in the boundary, so it is hard for our model to segment some extremely tiny objects very precisely.

\subsection{Qualitative Results on Extremely Long Video Clips}
\begin{figure*}[t]
	\centering
	\includegraphics[width=\textwidth]{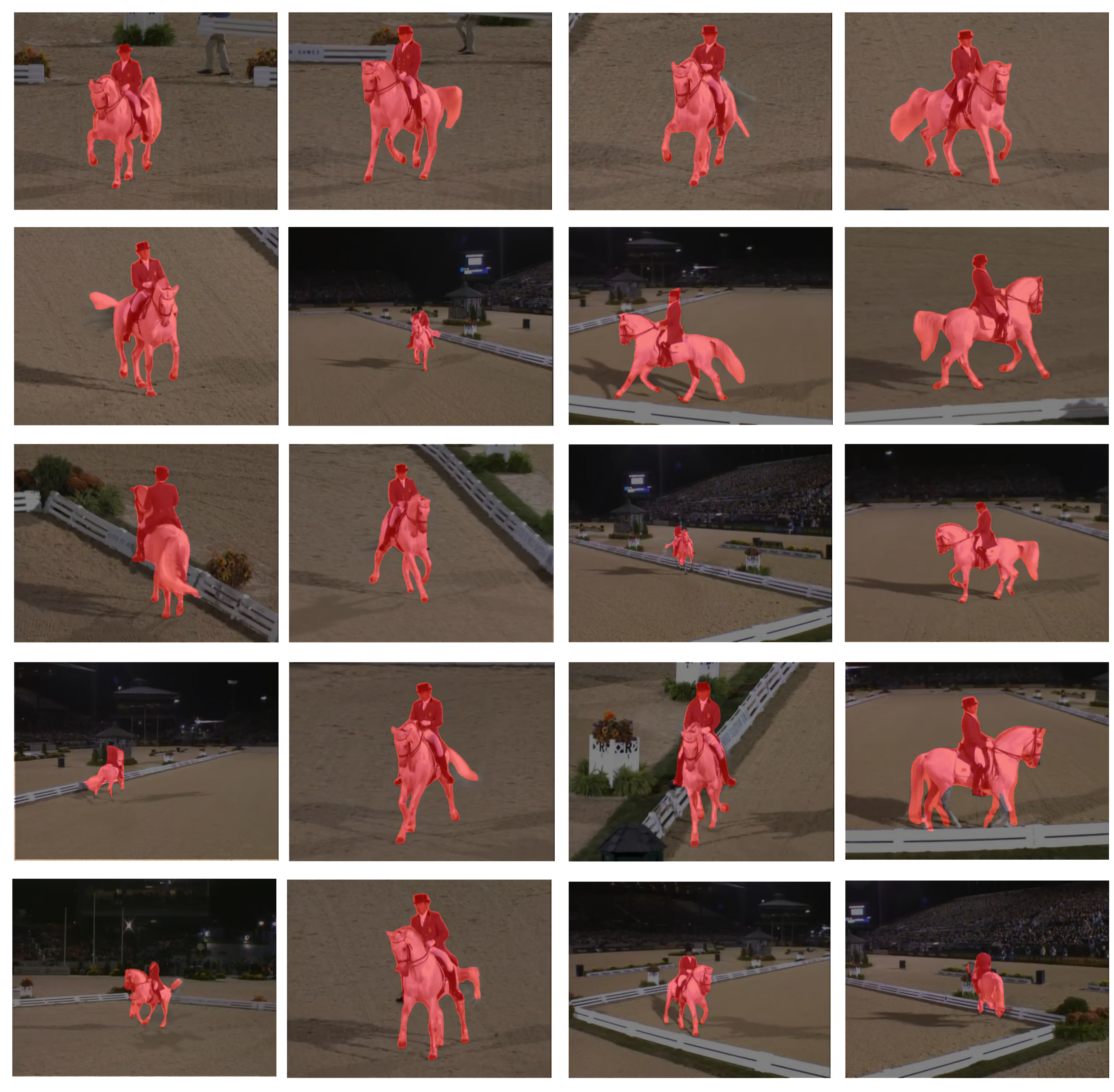}
\caption{We show a long-video case (\textit{dressage}) of \textit{a man riding a horse} with 3589 frames from \cite{liang2020video} and we sample 20 frames overtime here for illustration. The first frame with its mask (in the top left corner) is given as the reference for mask propagation, the propagated masks along the time are arranged from left to right and from top to bottom. Even if our model (Ours-Base) only leverages the guidance from the first and the previous frame, it can handle such cases with an extremely large number of video clips and large appearance changes well. Please zoom in on the figure to view it better.}
	\label{fig:long_video}
\end{figure*}
\begin{figure*}[t]
	\centering
	\includegraphics[width=\textwidth]{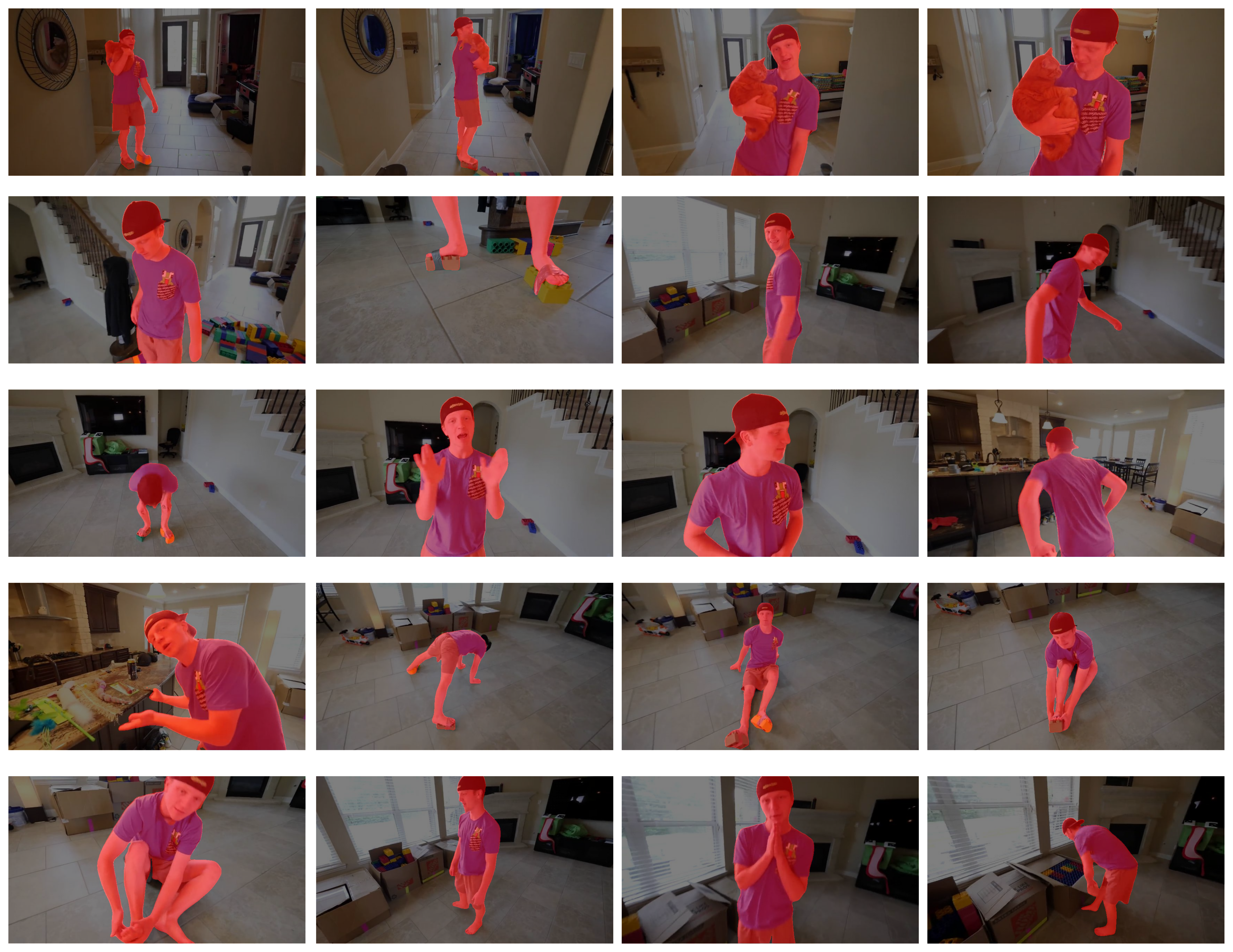}
\caption{We show a long-video case (\textit{blueboy}) of \textit{a man in a blue shirt who does a lot of actions} with 2406 frames from \cite{liang2020video} and we sample 20 frames over time for illustration. The first frame with its mask (in the top left corner) is given as the reference for mask propagation, the propagated masks along the time are arranged from left to right and from top to bottom. Even if our model (Ours-Base) only leverages the guidance from the first and the previous frame, it can handle such cases with an extremely large number of video clips and large appearance changes well. Please zoom in on the figure to view it better.}
	\label{fig:long_video_boy}
\end{figure*}
Considering that the standard large-scale VOS benchmarks, including DAVIS \cite{perazzi2016benchmark} and YouTube-VOS \cite{xu2018youtube}, are tailored for evaluation on short-term video clips, we further evaluate our proposed model (Ours-Base) on two long-term video clips (\textit{blueboy} and \textit{dressage}) from \cite{liang2020video}. These videos contain more than 1000 video frames and include more diverse appearance changes. Specifically, the video clip \textit{blueboy} contains 2406 frames and the video clip \textit{dressage} contains 3589 frames. For the mask propagation in these two video sequences, We use our model (Ours-Base) which is trained only on the YouTube-VOS \cite{xu2018youtube} and only leverages the guidance from the reference (first) and the previous frame. 
Even if our model (Ours-Base) only maintains a constant memory bank of reference object proxies from the first frame and the previous frame, it can handle these cases with an extremely large number of video clips and large appearance changes well. The qualitative results of the mask predictions of our model are illustrated in Fig.~\ref{fig:long_video} and Fig.\ref{fig:long_video_boy}. We sample 20 frames here for illustration.

\subsection{More Qualitative Results on Robustness to Perturbations}
\begin{figure*}[t]
\centering
\includegraphics[width=\textwidth]{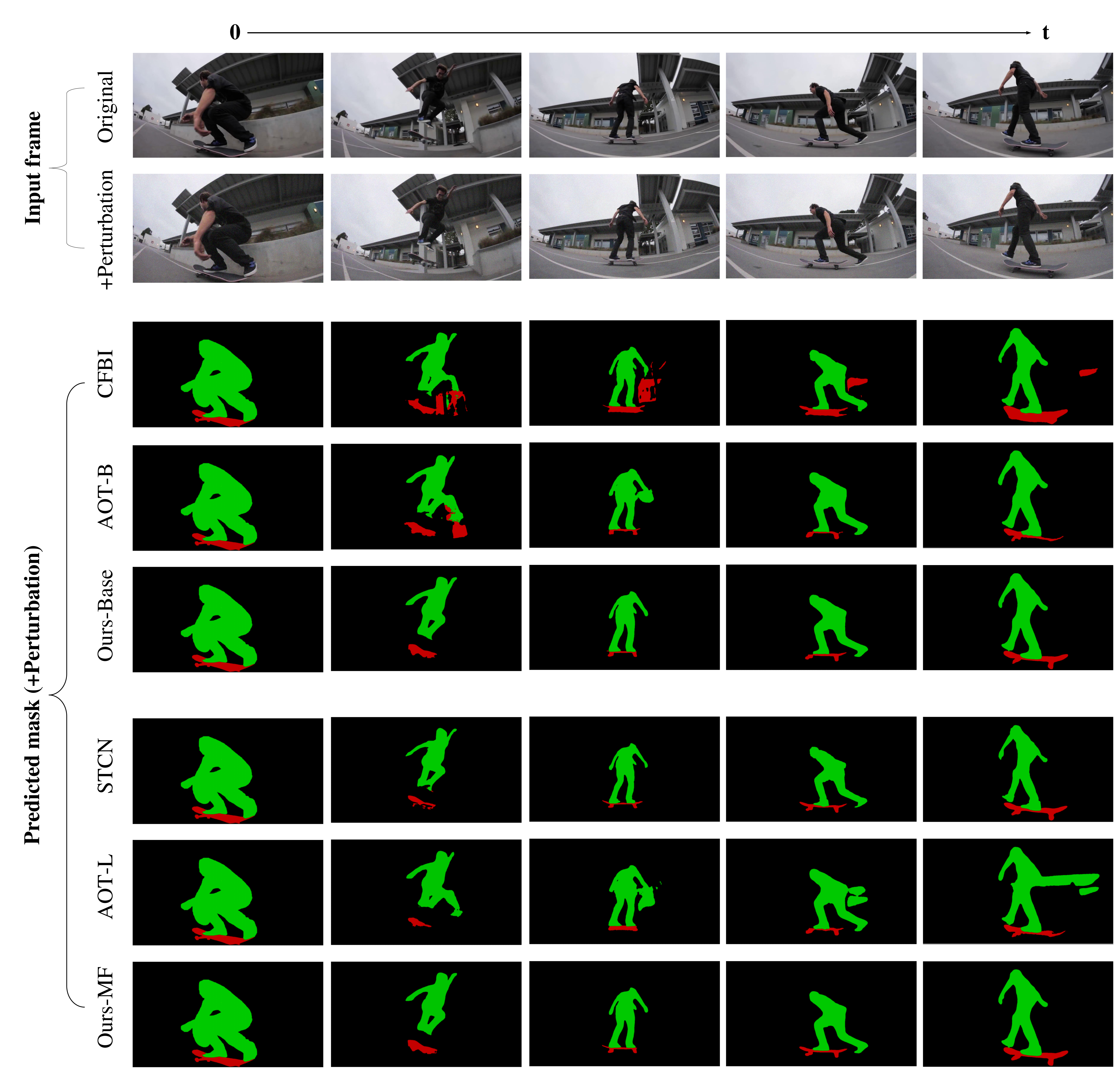}

\caption{A qualitative comparison between our model (Ours-Base, Ours-MF) and state-of-the-art models, including CFBI\cite{yang2020collaborative} (our baseline), AOT\cite{yang2021associating}, and STCN\cite{cheng2021rethinking}, on robustness against perturbations. This video clip is challenging due to the fast motion of the man playing the skateboard. Here, Gaussian noise is injected into the video as the perturbation. Our baseline model, CFBI \cite{yang2020collaborative}, can not handle such a case with fast motion and perturbations of Gaussian noise well. On the contrary, thanks to the proposed adaptive object proxy representation and discriminative object calibration, our model shows the best performance and robustness to perturbations among all the models. }
\label{fig:qualitative_robust1}                   
\end{figure*}

\begin{figure*}[t]
\centering
\includegraphics[width=\textwidth]{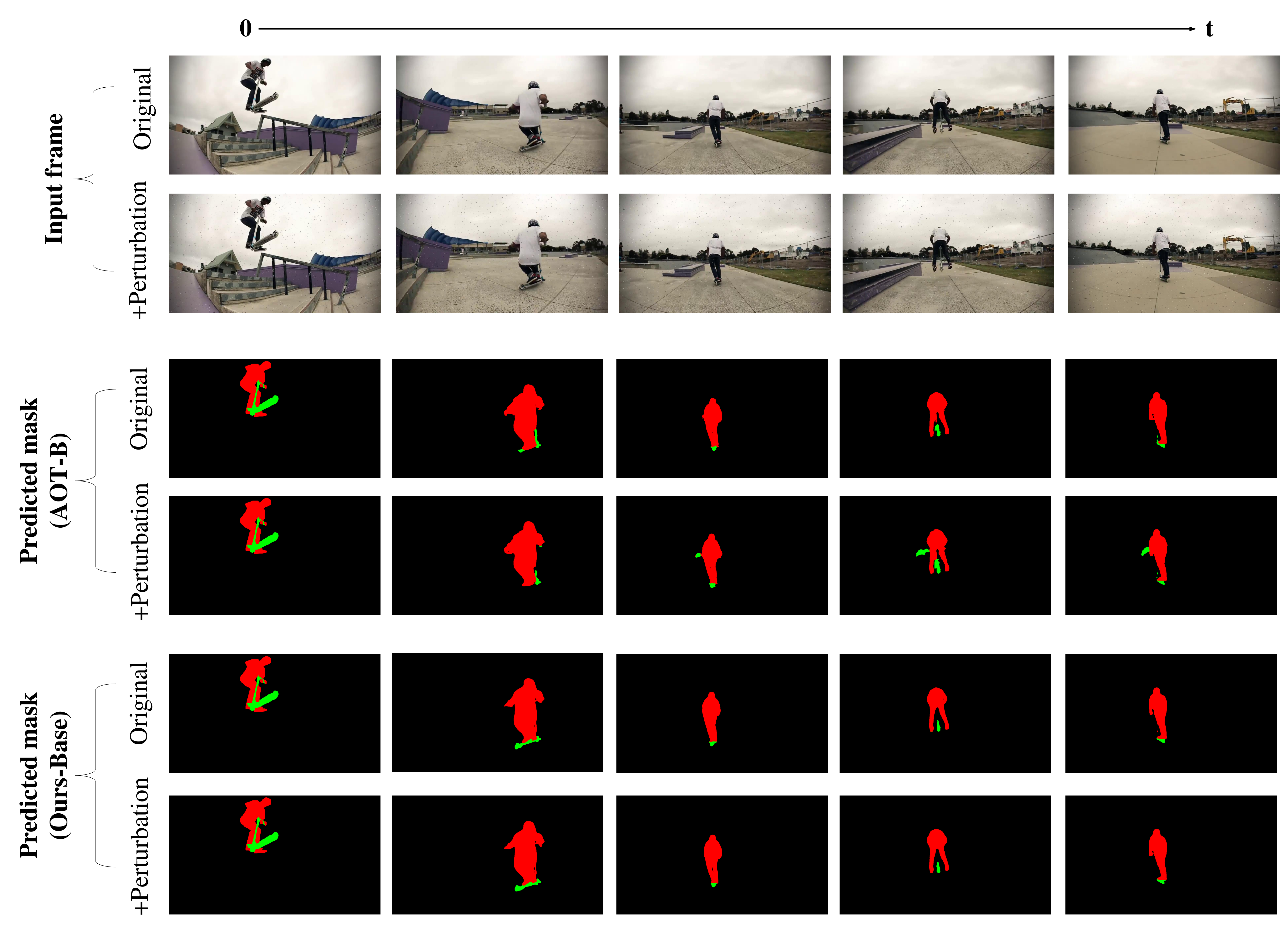}

\caption{A qualitative comparison between our model (Ours-Base) and a state-of-the-art model AOT-B \cite{yang2021associating} on robustness against perturbations of salt and pepper noise with 1$k$ points. Both two models perform well on the original clean video clip. However, our model outperforms AOT-B on perturbed videos with noise injected. }
\label{fig:qualitative_robust2}                   
\end{figure*}

\begin{figure*}[t]
\centering
\includegraphics[width=\textwidth]{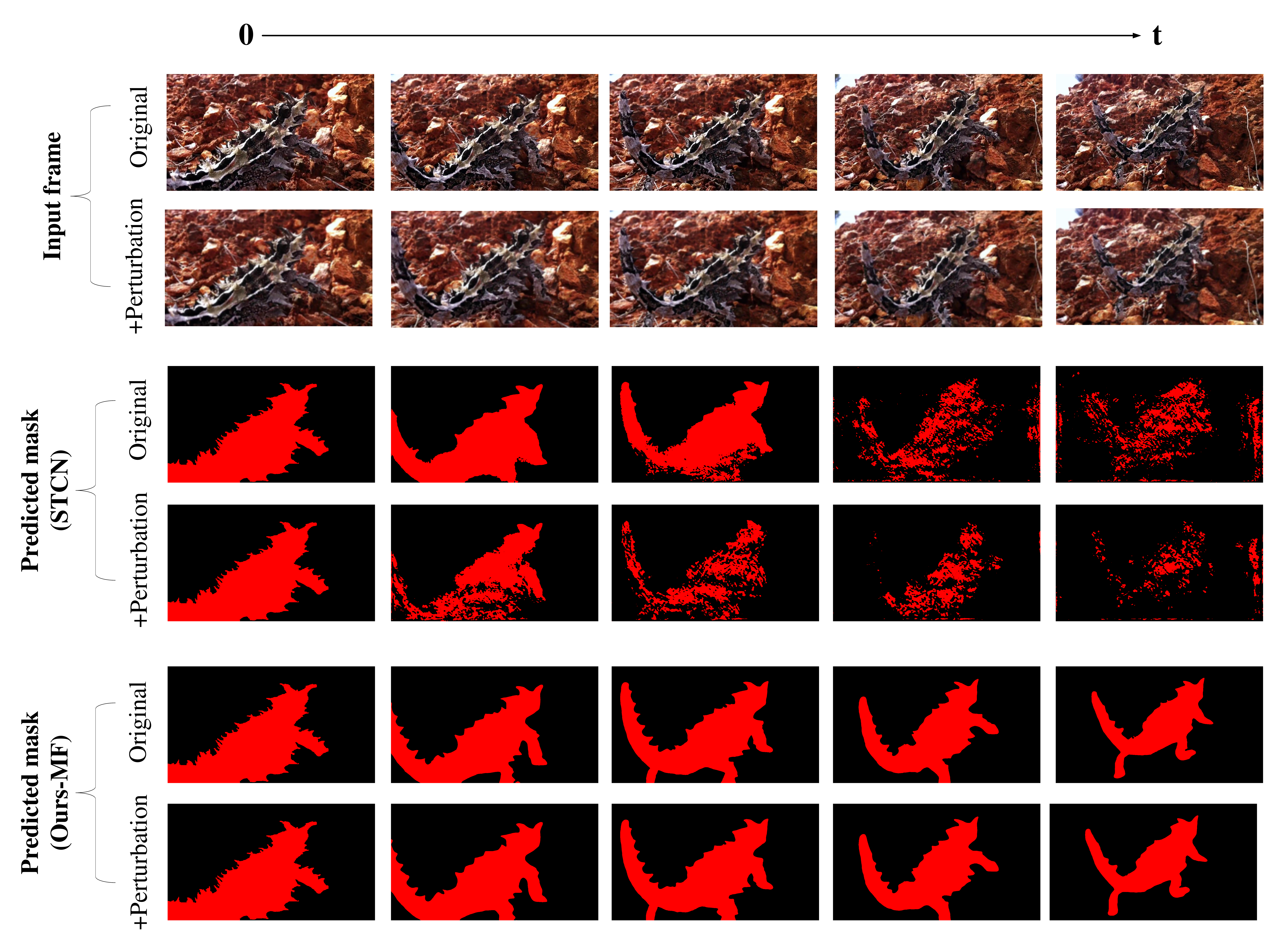}

\caption{A qualitative comparison between our model (Ours-MF) and a state-of-the-art model STCN \cite{cheng2021rethinking} on robustness against perturbations of Gaussian blur with $9\times9$ kernel. In this video clip, the target object to be segmented is partially similar to the background in texture. For mask prediction given the original clean video clip, the performance of STCN will degrade and become more and more unstable over time while our model can tackle this case well. For the prediction when the video clip is under perturbation, the performance degradation of STCN goes more violent. Contrary, our model shows stronger robustness to perturbations and there is no drop in performance under perturbation in such a video case.}
\label{fig:qualitative_robust3}                   
\end{figure*}

We show more qualitative cases to show the performance and the robustness of state-of-the-art models, including CFBI \cite{yang2020collaborative} (our baseline), AOT \cite{yang2021associating}, and STCN \cite{cheng2021rethinking}, and our model (Ours-Base and Ours-MF) when the input video clips are under perturbations. Concretely, we show three cases from YouTube-VOS \cite{xu2018youtube} 2019-version dataset with the perturbation of Gaussian noise, salt and pepper noise, and Gaussian blur in Fig.~\ref{fig:qualitative_robust1}, Fig.~\ref{fig:qualitative_robust2}, and Fig.~\ref{fig:qualitative_robust3} respectively. Our model shows stronger robustness to perturbations compared to other previous strong competitors in these challenging cases.

\subsection{More Cases for Adaptive Object Proxy Representation}
\begin{figure*}[t]
	\centering
	\includegraphics[width=\textwidth]{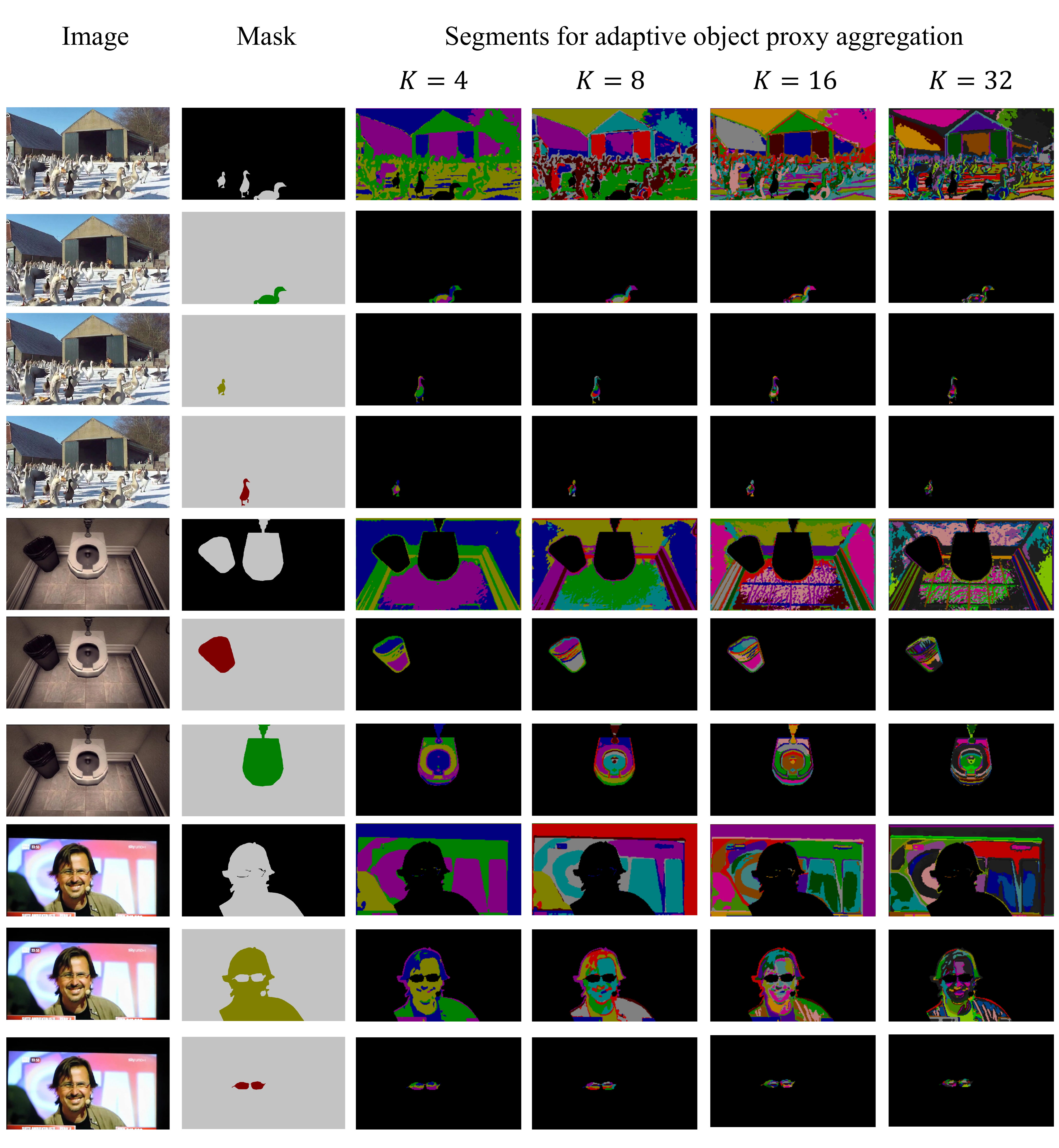}
\caption{Visualization of clustered $K$ segments implemented with $K$-means clustering for adaptive object proxy aggregation. Each segment in this figure represents a cluster to be aggregated as an adaptive object proxy during adaptive object proxy aggregation. The adaptive object proxies are further used to construct correspondence with the current frame feature.}
	\label{fig:clustering_appendix}
\end{figure*}

We provide more cases to illustrate the adaptive object proxy representation constructed with $K$-means clustering in Fig.~\ref{fig:clustering_appendix}. Each segment in Fig.~\ref{fig:clustering_appendix} represents a cluster to be aggregated as an adaptive object proxy. For the segments generated via clustering, we can find that regions of similar objects, the same object, or the spatially neighboring areas are clustered into the same group, which decomposes a complex region into several semantically-alike parts. Such an adaptive representation constructs a meaningful candidate pool for a query to do retrieval and calculate correlations for prototypical mask initialization. Meanwhile, due to the de-noising property of representation aggregation, our adaptive object proxies are more robust than the widely-used pixel-level proxies.

\subsection{More Qualitative Results on DAVIS}
\begin{figure*}[ht!]
	\centering
	\includegraphics[width=1\textwidth]{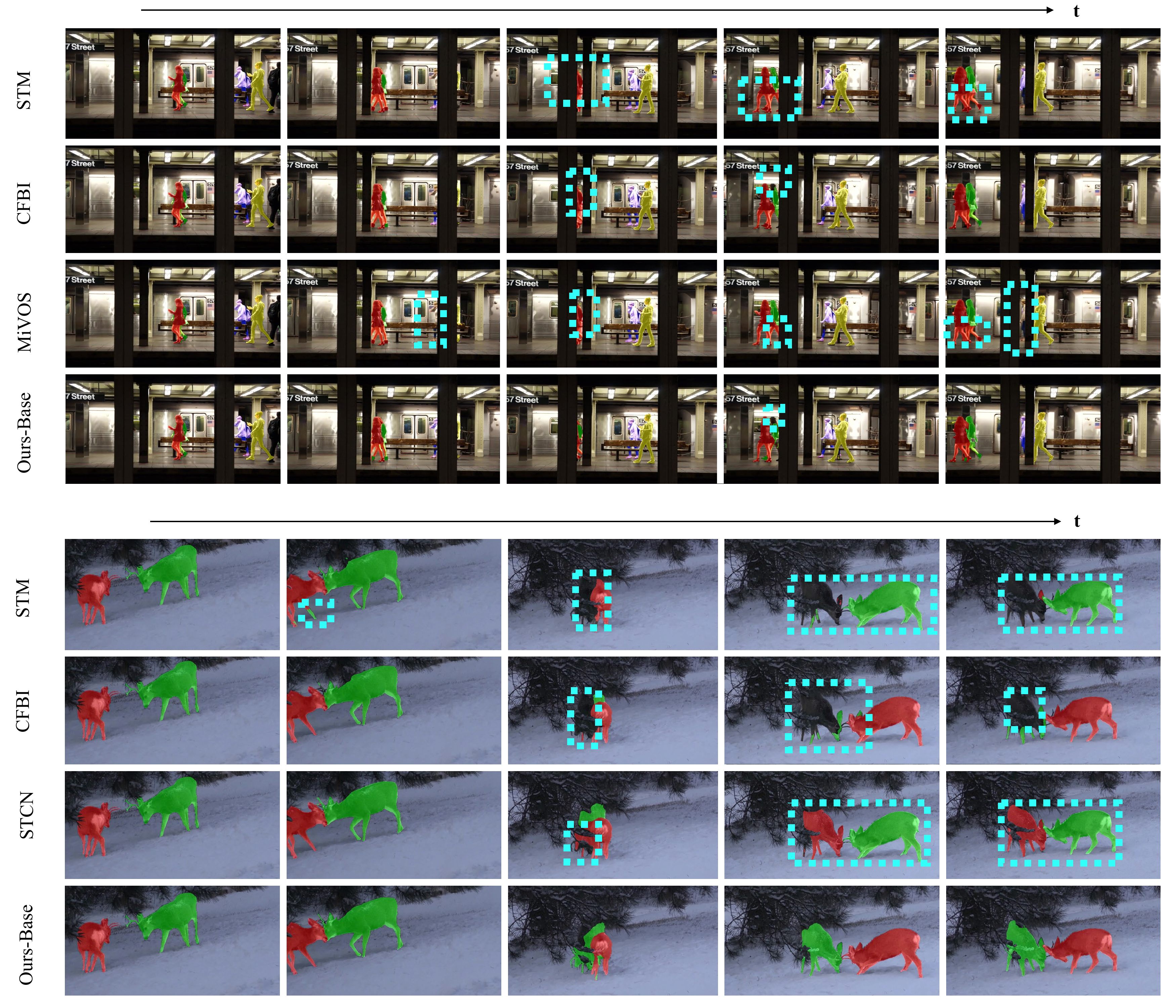}
\caption{Qualitative comparison to previous state-of-the-art methods, STCN \cite{cheng2021rethinking}, STM~\cite{oh2019video}, CFBI~\cite{yang2020collaborative} and MIVOS~\cite{cheng2021modular} on DAVIS17 ~\cite{ponttuset20182017} test-dev split. All of them are predicted with input resolution 480$p$. Our model (Ours-Base) shows stronger performance in these cases with multiple objects and object occlusions. Error regions are highlighted with light blue bounding boxes. Zoom in to view better.}
	\label{fig:qualitative_comparison_test_dev}
\end{figure*}

Fig.~\ref{fig:qualitative_comparison_test_dev} shows two hard cases of the 
    qualitative comparison to previous state-of-the-art methods (STM \cite{oh2019video}, CFBI \cite{yang2020collaborative} and MiVOS \cite{cheng2021modular}) on the challenging DAVIS17 ~\cite{ponttuset20182017} test-dev split. In the first case, although our model produces some minor prediction errors in the intermediate frames, errors are reduced in subsequent predictions. The second case shows that our model can handle cases with similar objects or occluded objects.



\section{More Implementation Details}
For data augmentations, we apply flipping, scaling, and balanced random-crop the same as \cite{yang2020collaborative}. Specifically, the balanced random crop is The input size for the model is $465 \times 465$.

For the training strategy, following \cite{oh2018fast,yang2020collaborative,yang2021collaborative}, the proposed network leverages the sequential training strategy where a clip of consecutive frames is sampled in each iteration. Concretely, a batch of video clips is sampled in each turn. For each video clip, we randomly sample a frame as the reference frame and a continuous $N + 1$ frames as the previous frame, and the current frame sequence with N frames. For the prediction of the reference frame, we use the ground-truth segmentation mask of the previous frame as the previous mask. For the prediction of the following frames, we use the latest prediction as the previous mask. 

For the construction of object proxy aggregation, we leverage $K-means$ clustering algorithm for implementation. We do not apply specific strategies for the initialization. 
The model is insensitive to the clustering algorithm and simple K-means clustering can already provide good performance.
The deviation of our final model for different initialization of clustering is small (about 0.1\% on YouTube-VOS\cite{xu2018youtube} 2019-version validation set).

\end{document}